\def\paperID{XXXXX}
\def\confName{CVPR}
\def\confYear{2026}

\def\paperTitle{Few-Step Diffusion Sampling Through Instance-Aware Discretizations}
\def\authorBlock{%
\begin{tabular}{c}
Liangyu Yuan\textsuperscript{1,2}\footnotemark[1]\thanks{~Equal contribution. $\dagger$ Corresponding author. This work was done during Liangyu Yuan's visit at WestLake University in 2025.}
\quad
Ruoyu Wang$^{1}$\footnotemark[1]
\quad
Tong Zhao$^{1,3}$\footnotemark[1]
\quad
Dingwen Fu$^{1}$
\\
Mingkun Lei$^{1}$
\quad
Beier Zhu$^{4}$
\quad
Chi Zhang$^{1\dagger}$
\\
$^{1}$Westlake University \quad $^{2}$Tongji University \\
$^{3}$Zhejiang University \quad
$^{4}$Nanyang Technological University \\
\end{tabular}
}

\newif\ifreview 
\newif\ifarxiv \newcommand{\arxiv}{\arxivtrue}
\newif\ifcamera 
\newif\ifrebuttal 

\def\eg{\emph{e.g.}} 

\def\ie{\emph{i.e.}}

\arxiv
\pdfoutput=1
\documentclass[10pt,twocolumn,letterpaper]{article}

\ifreview \usepackage[review]{cvpr} \fi
\ifarxiv \usepackage[pagenumbers]{cvpr} \fi
\ifrebuttal \usepackage[rebuttal]{cvpr} \fi
\ifcamera \usepackage{cvpr} \fi

\usepackage{xr-hyper}

\makeatletter
\newcommand*{\addFileDependency}[1]{
  \typeout{(#1)}
  \@addtofilelist{#1}
  \IfFileExists{#1}{}{\typeout{No file #1.}}
}

\makeatother
\newcommand*{\myexternaldocument}[1]{
    \externaldocument{#1}
    \addFileDependency{#1.tex}
    \addFileDependency{#1.aux}
}

\definecolor{cvprblue}{rgb}{0.21,0.49,0.74}
\usepackage[pagebackref,breaklinks,colorlinks,allcolors=cvprblue]{hyperref}
\usepackage[capitalize]{cleveref}
\crefname{section}{Sec.}{Secs.}
\crefname{table}{Table}{Tables}
\crefname{figure}{Fig.}{Figs.}

\ifarxiv \crefname{appendix}{App.}{Apps.}
\else \crefname{appendix}{Suppl.}{Suppls.} \fi

\usepackage{graphicx}	
\usepackage{amsmath}	
\usepackage{amssymb}	
\usepackage{booktabs}
\usepackage{times}
\usepackage{microtype}
\usepackage{epsfig}
\usepackage{caption}
\usepackage{float}
\usepackage{placeins}
\usepackage{color, colortbl}
\usepackage{stfloats}
\usepackage{enumitem}
\usepackage{tabularx}
\usepackage{xstring}
\usepackage{multirow}
\usepackage{xspace}
\usepackage{url}
\usepackage{subcaption}
\usepackage{xcolor}
\usepackage[hang,flushmargin]{footmisc}

\ifcamera \usepackage[accsupp]{axessibility} \fi

% ...

\ifarxiv  \fi

\newcommand{\R}[1]{{%
    \textbf{%
        \ifstrequal{#1}{1}{\textcolor{red}{R#1}}{%
        \ifstrequal{#1}{2}{\textcolor{blue}{R#1}}{%
        \ifstrequal{#1}{3}{\textcolor{magenta}{R#1}}{%
        \ifstrequal{#1}{4}{\textcolor{teal}{R#1}}{%
                           \textcolor{cyan}{R#1}%
        }}}}%
    }%
}}

\usepackage{graphicx}  
\usepackage{float}     
\usepackage{array}     
\usepackage{multirow}  
\usepackage{rotating}  
\usepackage{booktabs}  
\usepackage{caption}   
\usepackage{subcaption}
\usepackage{amsmath}
\usepackage{wrapfig}
\usepackage{cleveref}

\usepackage{algorithm}
\usepackage{algorithmic}

\newcommand{\toyNstep}{3}

\newcommand{\Ours}{\text{INDIS}}

\newcommand{\Studenttime}{\xi}
\newcommand{\Teachertime}{\psi}
\newcommand{\Priornet}{\phi}
\newcommand{\ODE}{\Psi}
\newcommand{\ErrGlobal}{\varepsilon_g}
\newcommand{\ErrInstance}{\varepsilon_o}
\newcommand{\ErrPN}{\varepsilon_i}

\unless\ifarxiv \myexternaldocument{_supplementary} \fi

\begin{document}
\title{\paperTitle}
\author{\authorBlock}
\maketitle


\begin{abstract}

Diffusion and flow matching models generate high-fidelity data by simulating paths defined by Ordinary or Stochastic Differential Equations (ODEs/SDEs), starting from a tractable prior distribution. The probability flow ODE formulation enables the use of advanced numerical solvers to accelerate sampling. Orthogonal yet vital to solver design is the discretization strategy. 
While early approaches employed handcrafted heuristics and recent methods adopt optimization-based techniques, most existing strategies enforce a globally shared timestep schedule across all samples. This uniform treatment fails to account for instance-specific complexity in the generative process, potentially limiting performance.
Motivated by controlled experiments on synthetic data, which reveals the suboptimality of global schedules under instance-specific dynamics, we propose an instance-aware discretization framework. Our method learns to adapt timestep allocations based on input-dependent priors, extending gradient-based discretization search to the conditional generative setting. 
Empirical results across diverse settings, including synthetic data, pixel-space diffusion, latent-space images and video flow matching models, demonstrate that our method consistently improves generation quality with marginal tuning cost compared to training and negligible inference overhead.

\end{abstract}

\section{Introduction}\label{sec:introduction}

Diffusion Probabilistic Models (DPMs)~\cite{ho2020denoising,song2021scorebased} and adjacent flow-matching models~\cite{albergo2023stochastic,lipman2023flow,liu2023flow} generate high-fidelity data by simulating trajectories defined by ODEs/SDEs, starting from a simple prior distribution (typically isotropic Gaussian). This iterative refinement process underpins their strong generative capabilities across diverse modalities~\cite{flux2024,li2022diffusion,vignac2023digress,liu2023audioldm,ehtesham2025movie}. However, the generative power comes at a price: the tedious sampling time required for high-quality generation.

Acceleration methods for diffusion models can be divided into two main groups, model distillation~\cite{salimans2022progressive,song2023consistency,liu2023flow} and training free acceleration~\cite{zhang2022DEISiPNDM,zhao2023unipc,lu2022dpm,liu2025timestep,ma2024deepcache}.
Model distillation enables extreme few-step generation but often leads to distillation cost comparable to training. Conversely, training-free methods avoid the heavy tuning cost with a trade-off on more steps. Among them, solver-based methods stand out as an architecture-agnostic choice, leveraging numerical ODEs/SDEs techniques for higher order, multistep sampling, providing more portable acceleration for pre-trained models.

\begin{figure*}[t] 
  \centering
  \includegraphics[width=1.0\textwidth]{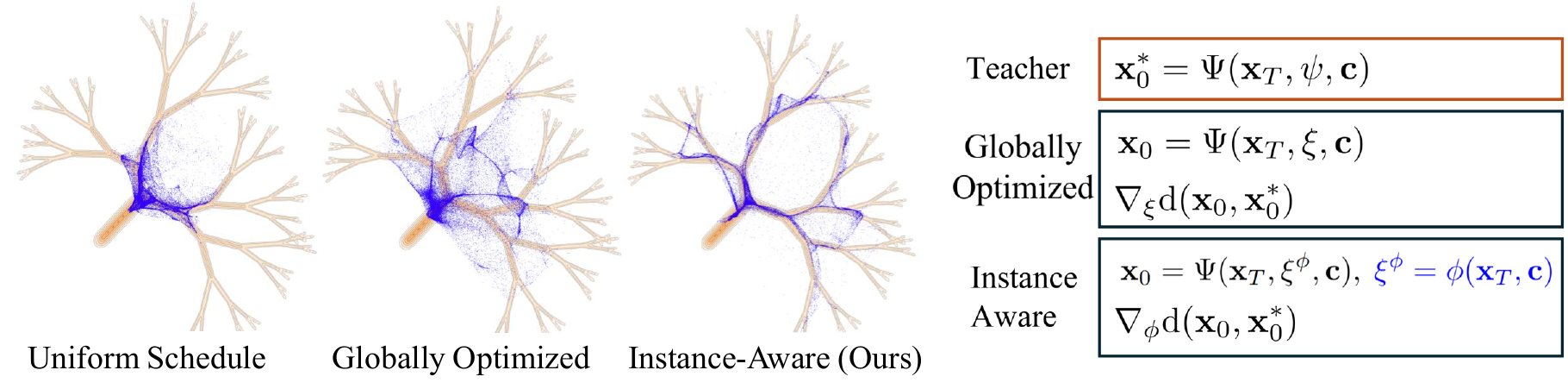}
\caption{Our effective instance-aware discretization improves sampling quality, by generating a tailored discretization $\xi^\phi$ for each initial noise $\mathbf{x}_T$ and condition $\mathbf{c}$, outperforming heuristic and globally optimized schedules. \textcolor{orange}{Orange} contour represents the ground truth data distribution, \textcolor{blue}{blue dots} represent the generated samples across different discretizations. ( \(\Psi(\cdot,\cdot,\cdot)\) represents the ODE path.)}
  \label{fig:teaser_data}
\end{figure*}

An essential aspect of solver design is the time discretization strategy. Initial approaches frequently relied on empirically derived heuristics, \eg uniform~\cite{ho2020denoising} or logSNR~\cite{lu2022dpm}. However, these heuristics were identified as suboptimal for maximizing efficiency. Subsequently, research efforts have increasingly focused on optimization-based techniques to search for better discretization strategies~\cite{tong2025LD3,kim2024distilling,park2025jump}.
Despite their improved efficiency, these optimization-driven approaches~\cite{tong2025LD3,chen2024GITS,xue2024accelerating,sabour2024align} share a critical limitation: they enforce a single, globally optimized timestep schedule for all starting priors. This design may neglect the intrinsic variability in data complexity across samples. 
In practice, different inputs can give rise to distinct sampling trajectories~\cite{ma2025inference,zhou2024golden}, each potentially benefiting from a different discretization.
To investigate this limitation, we first conduct a quantitative analysis using toy datasets (\Cref{fig:toy_gaussian}), revealing discernible performance gaps between globally optimized and instance-adaptive discretization schedules.
These observations highlight a critical limitation of the current discretization strategy and motivate the development of adaptive discretization strategies that dynamically allocate timesteps based on the characteristics of each input. 
Building on this insight,  we propose an effective method that generalizes previous gradient-based discretization search by taking the prior conditioning as input to produce instance-aware discretizations, as illustrated in~\Cref{fig:teaser_data} (right). 
To ensure scalability from synthetic analysis to high-dimensional image synthesis, we further introduce adaptation to handle conditional guidance and generalize the framework for alleviating mismatch issues commonly encountered in diffusion models literature. 
We name our algorithm \textit{\Ours} (\textit{Instance-Specific Discretization}).

Empirical results across diverse settings, including synthetic datasets, pixel space~\cite{Karras2022edm}, and latent space diffusion~\cite{rombach2022high} and flow matching models on images~\cite{flux2024} and videos~\cite{hacohen2024ltx}, validate the effectiveness of our approach. Moreover, our approach is a lightweight solver acceleration method with negligible tuning overhead versus training or distilling the base model, and marginal additional sampling cost. Our contribution can be summarized as follows:



\begin{itemize}
    \item We identify the limitations of global timestep discretization through synthetic experiments with quantitative analysis and propose an effective instance-aware discretization paradigm.
    \item We scale the paradigm of instance-aware discretization to high-dimensional image synthesis, by incorporating adaptations to manage conditional guidance and generalizing the framework to mitigate the exposure bias problem.
    \item Extensive experiments across diverse datasets and model types, including pixel-space diffusion, latent-space images and video flow matching models, validate the effectiveness of our approach.
\end{itemize}

\section{Related work}\label{sec:related work}

\noindent\textbf{Dedicated solvers for diffusion ODEs.}
Building upon the probability flow ODE formulation, significant research has aimed to accelerate diffusion sampling. DDIM~\cite{song2021ddim} pioneered this by using a non-Markovian process to reduce DDPM~\cite{ho2020denoising} steps from thousands to fewer than a hundred. Subsequently, DPM-Solver~\cite{lu2022dpm} and DPM-Solver++~\cite{lu2022dpm++} realize the semi-linear form of diffusion ODE and use knowledge from exponential integrators to improve sampling quality. PNDM~\cite{liu2022pseudo} and iPNDM~\cite{zhang2022DEISiPNDM} incorporated linear multistep methods for sampling efficiency. UniPC~\cite{zhao2023unipc} further proposed a unified predictor-corrector framework to minimize local truncation errors. RX~\cite{choi2025enhanced} uses Richardson Extrapolation to function as a plugin to improve sample quality across multistep methods.

\noindent\textbf{Diffusion solver fine-tuning}. Recognizing the potential limitations of applying fixed solver parameters derived from general numerical analysis, more recent research~\cite{zhou2024fast,shaul2024bespoke,shaul2024bespokenonstationary,kim2024distilling,zhu2025distilling,wang2025adaptivestochasticcoefficientsaccelerating,wang2025paralleldiffusionsolverresidual} has explored incorporating domain-specific information to enhance sampler performance. AMED~\cite{zhou2024fast} proposes to align the approximated mean value from low dimensionality across steps through tuning the intermediary time. Bespoke solver class (bespoke solver~\cite{shaul2024bespoke}, bespoke non-stationary solver~\cite{shaul2024bespokenonstationary}) in flow matching models proposes to learn the general form of the solver parameters given global/local supervision.

\noindent\textbf{Optimizing diffusion ODE timestep discretization.} Apart from dedicated solver design, timesteps discretization finetuning has recently garnered significant research interest. Various approaches~\cite{xue2024accelerating,sabour2024align,chen2024GITS,tong2025LD3} have been proposed, aiming to improve upon sub-optimal manually-crafted schedules. 
Both DMN~\cite{xue2024accelerating} and AYS~\cite{sabour2024align} formulate timestep selection as an optimization problem, solved using constrained trust-region methods and Monte Carlo sampling techniques, respectively.
GITS~\cite{chen2024GITS} leverages assumptions about the geometric regularity of the sampling trajectory, modeling discretization as a shortest path problem that minimizes accumulated local truncation errors. LD3~\cite{tong2025LD3} introduces a lightweight framework to explicitly learn the optimal discretization by minimizing the endpoint truncation error with respect to a teacher solver using gradient-based method. Optimization approaches for discretization also exist across other modalities~\cite{park2025jump}. 

\section{Preliminaries}\label{sec:preliminaries}

\subsection{Diffusion ODEs for sampling}
\label{subsec:diffusion_odes}

Diffusion Probabilistic Models (DPMs)~\cite{ho2020denoising,song2021scorebased,sohl2015deep} are generative models that learn to reverse a noising process. Data generation involves a learned score network, $s_{\theta}(\mathbf{x}_t,t)$, which approximates the score function $\nabla_{\mathbf{x}}\log q_t(\mathbf{x}_t)$ of the perturbed data density at time $t$. This score function can be equivalently parameterized via noise prediction $\epsilon_{\theta}(\mathbf{x}_t, t)$ or data prediction $\mathbf{x}_{\theta}(\mathbf{x}_t, t)$.

For sampling, a widely adopted approach is to use the deterministic probability flow ordinary differential equation (PF-ODE), whose trajectories share the same marginal densities as the reverse-time SDE:
\begin{align}
\begin{split}
\label{eq:probabilityflowODE}
    \mathrm{d}\mathbf{x}_t = \left[\mathbf{f}(t)\mathbf{x}_t - \frac{1}{2}\mathbf{g}^2(t)\nabla_{\mathbf{x}}\log q_t(\mathbf{x}_t)\right]\mathrm{d}t, \\ \mathbf{x}_T \sim \mathcal{N}(0,\mathbf{I}).
\end{split}
\end{align}

Here, the drift and diffusion-related coefficients $\mathbf{f}(t)=\frac{\dot{\alpha}_t}{\alpha_t}$ and $\mathbf{g}^2(t) =2\dot{\sigma}_t\sigma_t - 2 \frac{\dot{\alpha}_t}{\alpha_t} \sigma_t^2$ are determined by the forward process noise schedule $\alpha_t$ and $\sigma_t$. Common schedules are EDM-VE~\cite{Karras2022edm} (\(\alpha_t = 1,\sigma_t = t\)), Flow-Matching Optimal Transport~\cite{lipman2023flow,liu2023flow,albergo2023stochastic} (\(\alpha_t = 1-t,\sigma_t = t\)) Starting from a prior sample $\mathbf{x}_T\sim \mathcal{N}(\mathbf{0},\mathbf{I})$ (and incorporating conditional information $\mathbf{c}$ if available), integrating \Cref{eq:probabilityflowODE} from $t=T$ down to $t \approx 0$ using numerical ODE solvers forms the basis for efficient sample generation.

Building upon the ODE integration path in~\Cref{eq:probabilityflowODE} with noise prediction $\epsilon_{\theta}(\cdot,\cdot)$, various advanced solvers were designed to boost sampling~\cite{zhang2022DEISiPNDM,zhao2023unipc,lu2022dpm}. 
Given a predefined discretization schedule $\xi = \{\tau_k\}_{k=0}^N$ where $\tau_0 = t_0\to 0$ and $\tau_N = T$, the higher order multistep methods can be seen as approximations to the integrating ODE path. A generalized form can be expressed as:

\vspace{-0.50cm}
\begin{align}
\begin{split}
    \mathbf{x}_{k-1} = \mathcal{F}^k(\mathbf{x}_k,\Studenttime) &:= u_k\cdot \mathbf{x}_{k}  + \sum_{j=1}^M w_{k,j}\cdot\epsilon_{k+j}, \\ \epsilon_k &:= \epsilon_{\theta}(\mathbf{x}_{k},\tau_k).
\label{eq:linear_multistep_ODE}
\end{split}
\end{align}
\vspace{-0.35cm}

\(M\) is the order of steps, \(u_{k},w_{k,j}\) are the multistep coefficients dependent on the subset of discretization schedule \(\Studenttime\). We follow the notation of~\cite{tong2025LD3} and define the sampling endpoint as:

\vspace{-0.50cm}
\begin{align}
\begin{split}
\label{eq:ode_path}
    \mathbf{x}_0 &= \ODE(\mathbf{x}_T,\Studenttime) = \mathcal{F}_1 \circ \mathcal{F}_2 \cdots \mathcal{F}_N(\mathbf{x}_T,\Studenttime) \\
    &= \bar{u}_1\mathbf{x}_T + \sum_{j=1}^N \bar{w}_{j}\epsilon_j.
\end{split}
\end{align}
\vspace{-0.35cm}

Where \(\bar{u}_1 = \prod^{N}u_k\), \(\bar{w}_{j}\) is a linear combination of \(u_k,w_{k,j}\). Given a predefined network \(\epsilon_{\theta}\) and solver parameterization choice (\eg iPNDM~\cite{lu2022dpm}, DPM-Solver++~\cite{lu2022dpm++}), then the ODE path can be considered as a function of the initial value \(\mathbf{x}_T\) and timesteps \(\Studenttime\).

\subsection{Gradient based discretization search}\label{subsec:motives}

Traditional solver designs often rely on heuristic discretization schedules, such as Uniform $\Studenttime = \{\tau_i=\frac{i}{N}(T-t_0)+t_0\}$ or LogSNR $\Studenttime = \{\tau_i = \frac{i}{N}(\lambda_{T}-\lambda_{t_0})+\lambda_{t_0}\}$, where $\lambda_t =\log (\alpha_t/\sigma_t)$. Because these manually crafted heuristics are typically suboptimal, recent research has increasingly focused on optimizing discretizations~\cite{tong2025LD3,chen2024GITS,xue2024accelerating}.

Among these efforts, gradient-based search with endpoint error supervision has proven highly competitive, as it fundamentally accounts for both approximation and accumulated truncation errors. Specifically, let $\Studenttime$ and $\Teachertime$ denote the student and teacher discretization strategies, respectively, where $|\Teachertime| > |\Studenttime|$. The optimization objective is formulated as:
\begin{equation}
    \label{eq:my_general_ts_optim}
    \arg\min_{\Studenttime} \mathbb{E}_{\mathbf{x}_T \sim \mathcal{N}(0,\sigma_T^2\mathbf{I})} \left[ \mathrm{d}( \ODE(\mathbf{x}_T,\Teachertime),\ODE(\mathbf{x}_T,\Studenttime))\right],
\end{equation}
where $\mathrm{d}(\cdot,\cdot):\mathbb{R}^d\times \mathbb{R}^d\to \mathbb{R}$ represents a distance metric (e.g., MSE or LPIPS). Minimizing this objective is equivalent to optimizing the KL divergence between student and teacher samples in the data domain.

In LD3~\cite{tong2025LD3}, this is termed the ``hard optimization'' objective. To ease this, LD3 introduces a soft bound by treating the initial noise $\mathbf{x}_T$ as a learnable parameter, dynamically modifying noise-data pairs during training to reduce loss. We offer a new interpretation of this mechanism and compare it with our proposed method in~\Cref{subsec:discussion}.
\section{Method}\label{sec:benefits}
\subsection{Observations on toy examples}
\label{subsec:toy_example}
\begin{figure*}[t] 
    \centering 
    \includegraphics[width=1.0\textwidth]{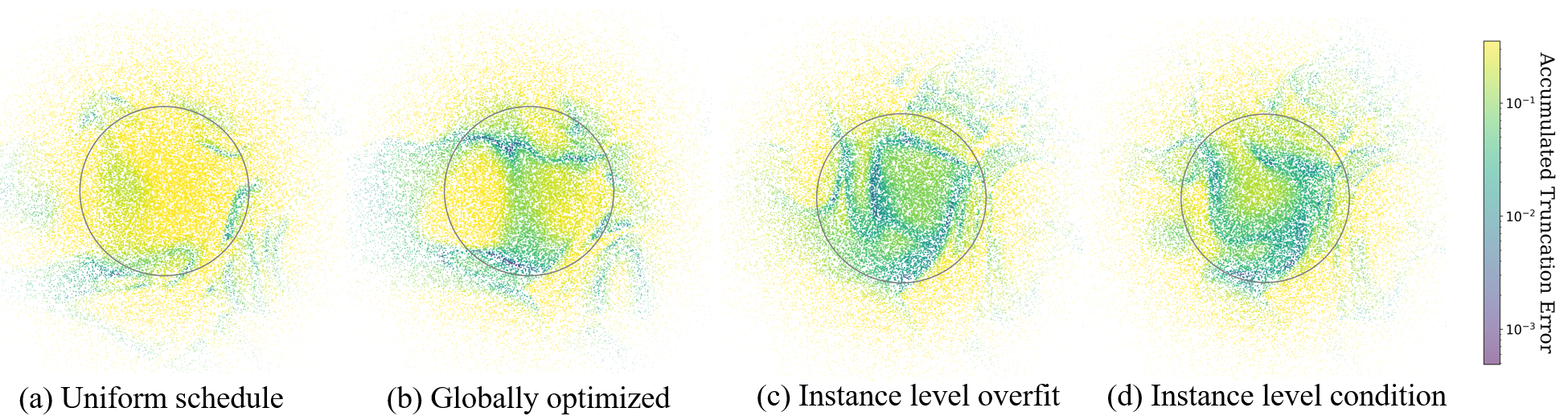}
    \caption{Comparison of endpoint(accumulated) errors for different timestep strategies (NFE=\toyNstep). Each point is an initial noise sample $\mathbf{x}_T\sim\mathcal {N}(0,\sigma_T^2\mathbf{I})$, colored by L2 error relative to 100-step euler as the ground truth. Methods: (a) Uniform timesteps. (b) Globally optimized timesteps. (c) Instance-specific timesteps (overfitted). (d) Instance-specific timesteps (learned through network \(\phi\)).}
    \label{fig:toy_gaussian}
\end{figure*}

Formally, recent learning-based methods~\cite{tong2025LD3,kim2024distilling} aim to optimize a single global set of timestep parameters, denoted as $\Studenttime^*$, which is uniformly applied across all initial samples $\mathbf{x}_T$. While this global optimization strategy can outperform fixed heuristic schedules, it inherently yields an optimal compromise only on average across all instances, rather than for each instance. In other words, if each sample were assigned its own optimal schedule, the global performance would naturally be optimal as well. But the reverse does not hold: a globally optimal schedule does not guarantee optimality at instance level.

This implies that the expected loss under a globally shared schedule, $\ErrGlobal$, serves as an upper bound on the expected loss achievable by an instance-specific approach, $\ErrPN$; ($\ErrGlobal \geq \ErrPN$). This asymmetry motivates the following research questions: \textbf{(RQ1)} \textit{To what extent can instance-specific discretization improve sampling performance compared to globally optimized schedule?} \textbf{(RQ2)} \textit{How can we effectively design a conditioning mechanism that produces a tailored timestep schedule for each instance?} To answer these questions, we conduct a set of controlled experiments on synthetic example designed to isolate and quantify the benefits of instance-level scheduling.

\noindent\textbf{Controlled experiment setup}. We consider the recursive tree branch data distribution from~\cite{Karras2024autoguidance} for better resemblance of the actual high-dimensional image data distribution, and change the noise schedule from VE to flow matching OT (\([T =80,t_0=0.002]\to [T=0.988,t_0=0.002]\)) for better trajectory and prior analysis. (Details of the toy are in the appendix).

We compare the following timestep strategies for sampling via the Euler method using $\toyNstep$ steps: \textbf{(a)} As a baseline, we employ uniform discretization, where timesteps are evenly spaced as $\tau_i = \frac{i}{N}(T - t_0) + t_0$. \textbf{(b)} Representing global optimization methods, we learn a single shared set of $\toyNstep$ timesteps by minimizing~\Cref{eq:my_general_ts_optim}, averaged over a training set of $20,000$ prior samples $\{ \mathbf{x}_T^i \}$. \textbf{(c)} As an oracle-style upper bound, we overfit a dedicated set of $\toyNstep$ timesteps for each individual prior sample $\mathbf{x}_T^i$, optimizing the error specific to that trajectory. For (b) and (c), we adopt MSE(Mean Square Error) as the distance metric $\mathrm{d}(\cdot, \cdot)$.

\noindent\textbf{Performance gap}. To ensure a fair comparison, (b) and (c) are trained and evaluated using the same set of prior samples $\{ \mathbf{x}_T^i \}$, which also serves as the sampling set in (a). The endpoint error is measured by comparing the final state of the sampled trajectory (using each strategy) to the ground truth obtained via 100-step Euler method from the same prior $\mathbf{x}_T^i$. As illustrated in~\Cref{fig:toy_gaussian}, strategy (c), which fits timesteps individually per sample, achieves significantly lower endpoint errors than both the uniform and globally optimized schedules, particularly in the high-density region near the center of the Gaussian prior distribution. Quantitatively, the instance-specific (overfitted) achieves an average MSE of $\ErrInstance = 0.0122$, representing a \textbf{50.2\%} reduction compared to the globally optimized schedule ($\ErrGlobal = 0.0245$). This discernible performance gap forms our \textbf{core motivation}: To realize effective instance-level discretizations, we propose to directly condition the timestep strategy on the starting point $\mathbf{x}_T$ of the deterministic ODE paths.

\noindent\textbf{Conditioning on prior}. Based on this motivation, we then train a lightweight network \(\Priornet(\cdot) : \mathbb{R}^d\to \mathbb{R}^{N}\) (where \(d\) is the noise/data dimension, \(N\) is the number of steps), taken the noise prior as input, and output instance level timesteps. We mark it as \textbf{(d)}. As illustrated in~\Cref{fig:toy_gaussian} (d), the conditioning is effective with regard to the global error estimation.

\begin{figure*}[t] 
    \centering 
    \includegraphics[width=1.0\textwidth]{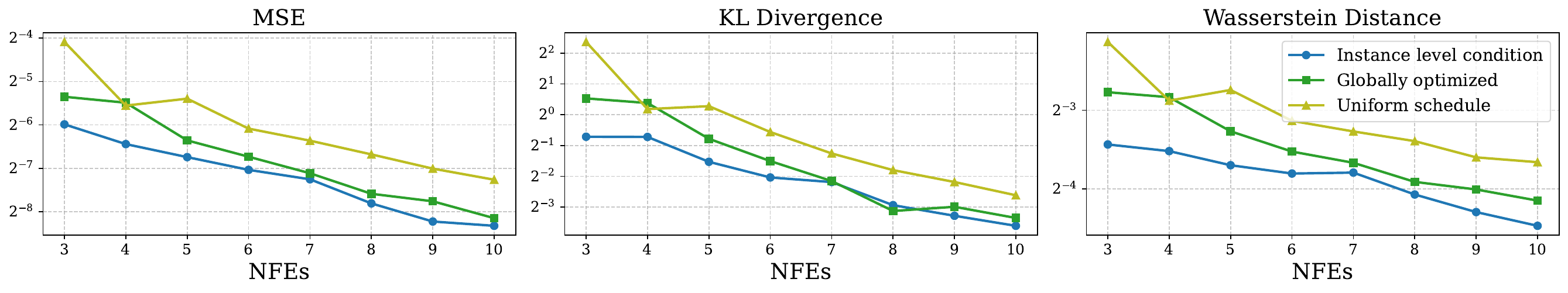}
    \caption{Quantitative comparison on synthetic experiments, evaluating MSE to teacher samples, KL divergence, and Wasserstein distance across various NFEs(log-scale). Methods include: uniform heuristics, globally optimized timesteps, and our proposed instance-level optimized timesteps conditioned on prior sample.}
    \vspace{-1.5em}
    \label{fig:quanttoy}
\end{figure*}

Quantitatively, we compare the instance-specific timestep generation (\(\Priornet\)) against globally optimized timestep schedules across various NFEs. The evaluation employs metrics assessing both average per-instance accuracy as MSE, and overall distributional similarity, using KL divergence and Wasserstein distance compared to ground truth distribution, to see how this instance-level correction contributes to distributional optimization objective.
As illustrated in Figure~\ref{fig:quanttoy}, the results consistently demonstrate that incorporating instance-level information translates into substantial global performance gains. Statistically, this contributes to \(25.86\%,21.85\%\) on KL divergence and Wasserstein distance. The effect of instance-level condition becomes more apparent in the low NFE regimes (3-6), increasing these contributions to an average of 45.67\% and 29.54\%. 

\vspace{-0.15cm}

\subsection{Scaling to discretization search for image synthesis}\label{subsec:method}

We now develop our practical method for high-dimensional image synthesis. We first detail two effective adaptations, then proceed to present the general framework.

\noindent\textbf{Incorporating conditional guidance}
\label{subsec:conditioning}. Many contemporary applications of diffusion models involve conditional generation, where the sampling process is guided by auxiliary condition $\mathbf{c}$, such as class labels~\cite{karras2024edm2,Karras2022edm} or text prompts~\cite{rombach2022high,flux2024}. This guidance mechanism actively influences the evolution of the state $\mathbf{x}_t$ along the ODE trajectory, alongside the initial noise sample $\mathbf{x}_T$. Our framework accommodates this by incorporating the conditional information $\mathbf{c}$ as an additional input to the network $\Priornet$ that predicts the timestep parameters \( \Studenttime = \Priornet (\mathbf{x}_T, \mathbf{c})\). 
Specifically, we consider two types of conditional guidance \(\mathbf{c}\), \ie, class labels and prompt embedding. Details of the implemented network architecture can be referred to~\Cref{fig:network} and in the appendix.

\noindent\textbf{Time and scale shift factors}. 
The noise end state of training diffusion model often contains a minority part of data information (Take VE EDM for example, \(\mathbf{x}_T = \mathbf{x}_0 + \sigma, \sigma\sim \mathcal{N}(0,\sigma_T\mathbf{I})\)), while during sampling we always start from isotropic Gaussian \(\mathbf{x}_T\sim \mathcal{N}(0,\sigma_T\mathbf{I})\)) This is often noted as the \textbf{exposure bias/mismatch} problem in diffusion models~\cite{lin2024common,ning2024elucidating,li2024alleviating}. 
Taking this into consideration, we integrate potential correction factors (time input shifts and output scaling) as learnable parameters within our instance-aware optimization framework, rather than employing shared heuristics or statistically dependent parameters~\cite{li2024alleviating} used in prior work. Therefore, we adhere the setting in~\cite{zhou2024fast,tong2025LD3} to generalize the framework for alleviating the bias problem. Specifically, we reframe the transformed function evaluation \(\hat{\epsilon}_{\theta}\) as follows:
\begin{align}
\begin{split}
    \hat{\epsilon}_{\theta}(\mathbf{x}_n,\tau_n,\Delta \tau_n, \gamma_n) := \gamma_n\cdot\epsilon_{\theta}(\mathbf{x}_n,\tau_n+\Delta \tau_n), \\
    \Studenttime^{\Priornet} = \{\tau_n,\Delta\tau_n, \gamma_n\}_{n=1}^N = \phi(\mathbf{x}_T, \mathbf{c}).
\label{eq:instanceshiftfactors}
\end{split}
\end{align}
Where we incorporate \(\tau\) and \(\Delta \tau\) to make temporal and spatial shift, in an attempt to alleviate exposure bias. We define \(\{\tau_n,\Delta\tau_n, \gamma_n\}_{n=1}^N\) as the general discretization set, deciding the timesteps along with the function calls, applicable to be combined with existing solver parameterization. 
Here we present practical implementation of the network output \(\Priornet\), given \(N\) step, we first output the instance factors:
\begin{align}
\begin{split}
    O &= \Priornet(\mathbf{x}_T,\mathbf{c}),~ O = [o_\tau,o_{\Delta\tau}, o_{\gamma}]^T\in \mathbb{R}^{3\times N}\\
    \Delta{\tau} &= b_{\Delta\tau}\cdot \tanh{(o_{\Delta\tau}/2)},\\ \gamma &= b_{\gamma}\cdot \tanh{(o_{\gamma}/2)}+1,
\end{split}
\end{align}
Then we apply the softmax parameterization along with~\Cref{eq:instanceshiftfactors} to obtain the final discretizations.
\vspace{-0.2cm}
\begin{align}
\begin{split}
    \tau &= \frac{f(\cdot)-f(0)}{f(N)-f(0)}\cdot(T-t_0) + t_0 \\
    &\text{where }. f(i) = \sum_{n=i}^N\text{softmax}(o_{\tau})[n]. 
\end{split}
\end{align}
Where \(b_{\Delta\tau},b_{\gamma}\) is the predefined bounding parameter to ensure stability. For parameterization of the main timestep \(\{\tau_{i}\}_{i=1}^N\), we adhere to the setting in LD3 to ensure monotonicity. Further details are provided in the appendix.

\subsection{The proposed INDIS method}
Building upon synthetic analysis and tailored adaptation, here we present the implementation of instance-aware discretization \(\Ours\) in a formulated way. Given the initial sampling point \(\mathbf{x}_T \in\mathbb{R}^d\) with conditional information \(\mathbf{c} \in\mathbb{R}^e\) available, we design a network 
\(\Priornet(\cdot,\cdot) : \mathbb{R}^d\times \mathbb{R}^e\to \mathbb{R}^{3\times N}\), 
taking prior conditioning as input, then output the tailored discretizations. 
Given the teacher discretization strategy \(\Teachertime\), the optimization objective can be defined as:
\begin{align}
\begin{split}
\label{eq:my_general_ts_optim_instance}
    \arg\min_{\Priornet} \mathbb{E}_{\mathbf{c}\sim \mathcal{C}, \mathbf{x}_T \sim \mathcal{N}(0,\sigma_T^2\mathbf{I})} \left[ \mathrm{d}( \ODE(\mathbf{x}_T,\Teachertime, \mathbf{c}),\ODE(\mathbf{x}_T,\Studenttime^{\Priornet}, \mathbf{c}))\right].
\end{split}
\end{align}
Where \(\mathcal{C}\) is the set of conditions (\eg class labels, text prompts). 
We then present our discretization training pipeline.
$\Ours$ can be integrated into various differentiable ODE solvers for improved discretization search and with negligible additional computational overhead.

\begin{algorithm}[H]
\caption{Tuning \Ours}
\label{alg:training}
    \begin{algorithmic}[1] 
        \STATE Solver parameterization \(\Psi(\cdot,\cdot)\). Condition Set $\mathcal{C}$ if available, teacher discretization $\Teachertime$, prior conditioning net \(\Priornet\)
        \STATE Dataset: $\mathcal{D}\leftarrow \{\mathbf{c}\sim \mathcal{C},~\mathbf{x}_T\sim \mathcal{N}(0,\sigma_T\mathbf{I}),~ \mathbf{x}_0^* =\ODE(\mathbf{x}_T,\Teachertime,\mathbf{c})\}$.\textcolor{blue}{\COMMENT{$\text{Data Preparation}$}}
        \REPEAT
            \STATE Sample $\mathbf{x}_T, \mathbf{c},\mathbf{x}_0^*$ from $\mathcal{D}$
            \STATE $\Studenttime^{\phi} =\{\tau_n,\Delta\tau_n, \gamma_n\}_{n=1}^N = \phi (\mathbf{x}_T,\mathbf{c})$
            \STATE
            \textcolor{blue}{\COMMENT{$\text{Forward pass of  prior conditioning network}$}}
            \STATE $\mathbf{x}_0 = \ODE(\mathbf{x}_T,\Studenttime^{\phi},\mathbf{c})$ \textcolor{blue}{\COMMENT{$\hat{\epsilon}_n =  \gamma_n\cdot\epsilon_{\theta}(\mathbf{x}_n,\tau_n+\Delta \tau_n)$}}
        \STATE Take gradient step: $\nabla_{\Priornet} \mathrm{d}(\mathbf{x}_0, \mathbf{x}_0^*)$
        \UNTIL{convergence}
    \end{algorithmic}
\end{algorithm}

\vspace{1.5ex}
\hfill

\begin{figure}[ht] 
    \centering 
    \includegraphics[width=0.48\textwidth]{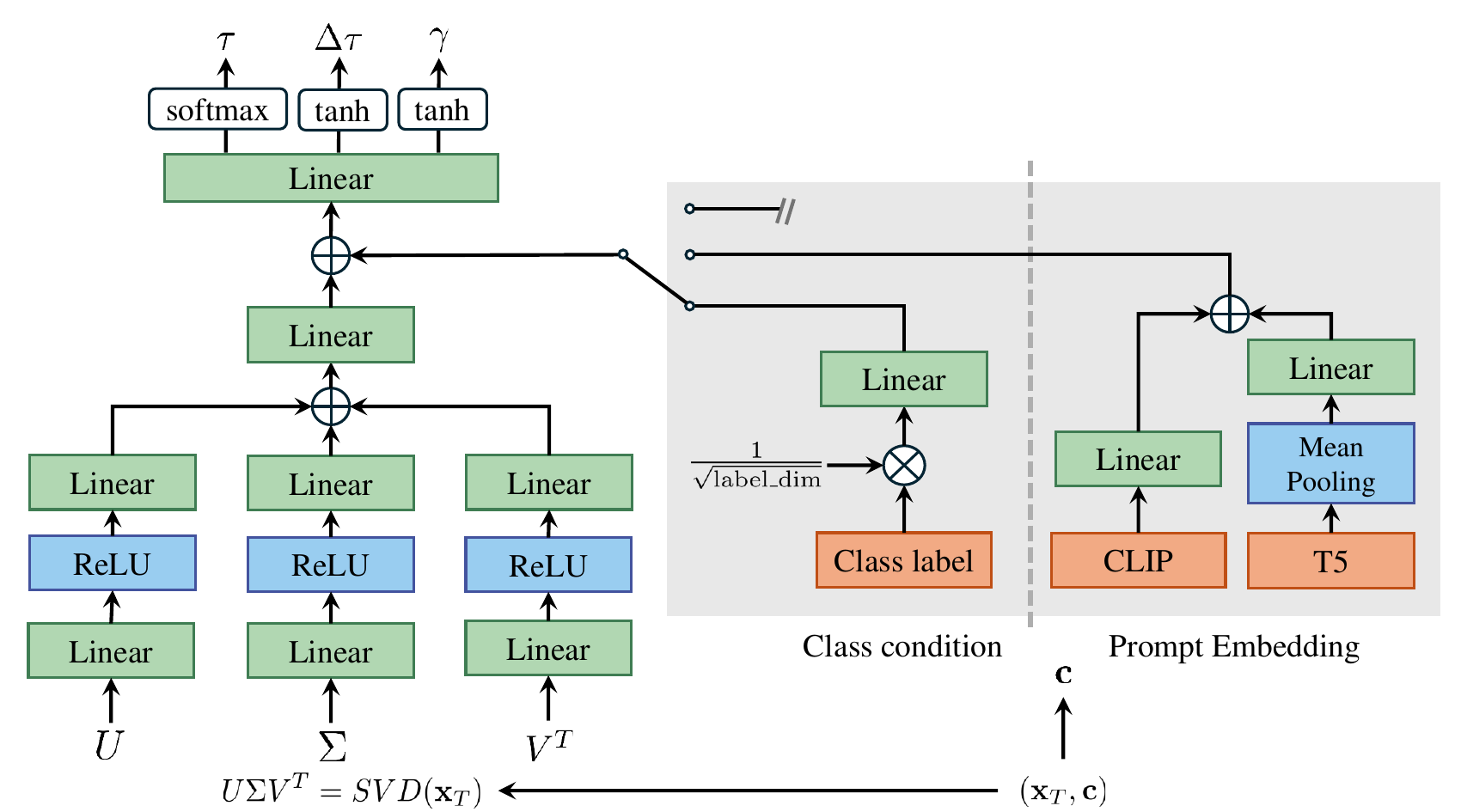}
    \caption{Architectural design of the proposed lightweight prior conditioning network. When conditional information is available, class indices are first scaled by a factor of $\frac{1}{\sqrt{\text{label\_dim}}}$ and then processed through a linear layer. For prompt embeddings (FLUX.1-dev), T5 embeddings undergo mean pooling to reduce dimensionality before being concatenated with CLIP embeddings.}
    \label{fig:network}
    \vspace{-0.7cm}
\end{figure}

\begin{table*}[b] 

\centering
\label{tab:pixel_fid}
\resizebox{1.0\textwidth}{!}{
\renewcommand{\arraystretch}{1.2}
\scriptsize
\begin{tabular}{@{}l@{\hskip 6pt}c@{\hskip 3pt}c@{\hskip 3pt}c@{\hskip 6pt}c@{\hskip 6pt}c@{\hskip 3pt}c@{\hskip 3pt}c@{\hskip 6pt}c@{\hskip 6pt}c@{\hskip 3pt}c@{\hskip 3pt}c@{\hskip 6pt}c@{\hskip 6pt}c@{\hskip 3pt}c@{\hskip 3pt}c@{}}
\toprule
& \multicolumn{3}{c}{CIFAR10 32$\times$32} & & \multicolumn{3}{c}{FFHQ 64$\times$64} & & \multicolumn{3}{c}{AFHQv2 64$\times$64} & & \multicolumn{3}{c}{ImageNet 64$\times$64} \\
\cmidrule(r){2-4} \cmidrule(r){6-8} \cmidrule(r){10-12} \cmidrule(){14-16}
Method & NFE=3 & NFE=5 & NFE=7 & & NFE=3 & NFE=5 & NFE=7 & & NFE=3 & NFE=5 & NFE=7 & & NFE=3 & NFE=5 & NFE=7 \\
\midrule
Best Heu. & 57.39 & 17.35 & 7.61 && 72.29 & 17.52 & 8.76 && 40.24 & 9.01 & 4.73 && 44.93 & 15.53 & 8.64 \\
DMN        & 77.69 & 12.93 & 5.42 && 178.09& 20.93 & 10.17&& 178.76& 26.11 & 13.03&& 33.72 & 10.47 & 5.39 \\
AMED       & 18.49 & 7.59 & 4.36 && 26.87 & 12.49 & 6.64 && 31.82 & 7.34 & 4.39 && 28.06 & 10.74 & 6.66 \\
GITS       & 25.98 & 6.77  & 3.43 && 26.41 & 8.85  & 5.36 && 24.17 & 8.72  & 5.48 && 26.41 & 9.85  & 6.44 \\
LD3 & 16.52 & 5.32  & 3.37 && 23.86 & 8.56  & 4.69 && 17.94 & 6.09  & 2.77 && 27.82 & 11.55 & 5.63 \\
\midrule
\textbf{\Ours} & \textbf{9.26} & \textbf{3.31} & \textbf{2.60} && \textbf{17.72} & \textbf{6.91} & \textbf{3.90} && \textbf{10.15} & \textbf{3.48} & \textbf{2.37} && \textbf{18.96} & \textbf{7.28} & \textbf{4.94} \\
\bottomrule
\end{tabular}%
}
\captionsetup{width=1.0\textwidth} 
\vspace{0.15cm}

\caption{FID Comparison for pixel-space DPMs on CIFAR10, FFHQ, AFHQv2 and class conditional ImageNet64, reporting for NFE=3, 5, and 7.  Best Heu. represent the best heuristics schedule among uniform, logSNR and polynomial schedules. The complete results are provided in the appendix.}
\end{table*}

\vspace{-1.0cm}
\noindent\textbf{Implementation details.}
As illustrated in~\Cref{alg:training}, our training process commences with the generation of a dataset comprising tuples of $(\mathbf{c}, \mathbf{x}_T, \mathbf{x}_0^*)$, where $\mathbf{x}_T \sim \mathcal{N}(0, \sigma_T^2\mathbf{I})$ is the initial noise, $\mathbf{c}$ represents optional conditional information, and $\mathbf{x}_0^*$ is the corresponding target endpoint pre-computed using a higher-NFE teacher solver with heuristic discretization. Various multistep methods (\eg DPM-Solver, UniPC, iPNDM) were considered for both teacher and student roles, we empirically found iPNDM to yield superior performance and thus selected it as the base solver structure for both. We store the random generator state instead of raw noise, thus the memory cost of this part is negligible.

During each training iteration, given an initial noise $\mathbf{x}_T$ and $\mathbf{c}$ from a batch, our parameter prediction network $\Priornet(\cdot,\cdot)$ first calculates the instance-specific discretization parameters $\Studenttime$. Subsequently, ODE sampling is performed using these tailored parameters to generate a student sample $\mathbf{x}_0$. Consistent with practices in~\cite{tong2025LD3,song2023consistency}, we employ the LPIPS~\cite{zhang2018unreasonable} as the distance metric $\mathrm{d}(\cdot,\cdot)$ in pixel domain between $\mathbf{x}_0$ and the cached teacher target $\mathbf{x}_0^*$. The parameters of the network $\Priornet$ are then updated via gradient-based optimization using Adam with cosine learning rate schedule. Further details on training implementation are provided in appendix.

At inference, generating the instance-specific parameters requires a single forward pass of \(\Priornet(\cdot,\cdot) : \mathbb{R}^d\times \mathbb{R}^e\to \mathbb{R}^{3\times N}\). This introduces minimal computational overhead compared to the total $N$ evaluations of the main diffusion model $\epsilon_\theta$. Formally, the sampling requires an extra forward pass of the lightweight network \(\Priornet\) to get instance-level discretizations:

\begin{align}
\begin{split}
    \mathbf{x}_0 = \ODE(\mathbf{x}_T,\Studenttime^{\phi},\mathbf{c}), \quad \Studenttime^{\phi} =  \phi (\mathbf{x}_T,\mathbf{c})\\ \mathbf{x}_T\sim \mathcal{N}(0,\sigma_T\mathbf{I}), ~\mathbf{c}\sim \mathcal{C}
\end{split}
\end{align}

\noindent\textbf{Efficiency analysis}. The forward pass of our prior condition network is negligible to the function calls of diffusion model.
For example, on 5-NFE setting, this overhead constitutes $2.5\%$ of the total sampling time for CIFAR10 and $2.3\%$ on FLUX.1-dev. Details could be referred in the appendix.


\noindent\textbf{Discussion.}
\label{subsec:discussion} The impact of initial noise of diffusion sampling has been extensively studied in recent literature~\cite{eyring2024reno,ma2025inference,zhou2024golden}, indicating that some noise are better than others. 
LD3 takes noise \(\mathbf{x}_T\) as a learnable parameter, we hypothesize that the noise is updated to get closer to better noise that result more closely to the data manifold. Then LD3 use a single set of hyperparameter timesteps to better align with those more important noise. This will help improve convergence speed and improve quality, but might quickly reach performance plateau when scaling up the dataset (e.g. \(> 100\)), due to the error between the original and updated noise given fixed image. 
Our work improves by directly assigning each prior \(\mathbf{x}_T\) a tailored discretization, giving the discretization more expressive power, thus shows better empirical performance when scaling up the dataset to thousands.

\vspace{-1.5ex}

\section{Experiments}\label{sec:experiments}

\subsection{Setup}

\label{subsec:experimental_setup}
\noindent\textbf{Pretrained models.} We use established pretrained diffusion and flow matching models for pixel-space and latent-space generation tasks.
For pixel-space diffusion models, we adopt the official EDM~\cite{Karras2022edm} pretrained checkpoints for CIFAR-10 ($32 \times 32$)~\cite{krizhevsky2009learning}, ImageNet ($64 \times 64$)~\cite{russakovsky2015ImageNet}, FFHQ ($64 \times 64$)~\cite{karras2019style}, and AFHQv2~\cite{choi2020stargan} ($64 \times 64$).
For latent-space text-to-image generation, we employ Stable Diffusion~\cite{rombach2022high} checkpoints for LSUN-bedroom and the guidance-distilled version of Flux (FLUX.1-dev)~\cite{flux2024}. For videos, we use LTX-Video~\cite{hacohen2024ltx}.

\noindent\textbf{Baseline methods.}  
We compare our method against recent open-source discretization techniques for diffusion models, specifically DMN~\cite{xue2024accelerating}, GITS~\cite{chen2024GITS} and LD3~\cite{tong2025LD3}. We also include AMED~\cite{zhou2024fast} to validate the effectiveness of our instance-aware approach. Since AFS~\cite{dockhorn2022genie} can be considered an orthogonal strategy that saves one NFE compared to discretization (in small scale dataset), we report the best result from with and without AFS for fair comparison (for AMED, we apply AFS to odd NFEs). For FlUX.1-dev and LTX-video, we don't use AFS. We also include results from the best-performing heuristic schedules among previously proposed manually-crafted options based on LogSNR, Uniform, and polynomial schedules. We keep our solver choice consistent with iPNDM, we report the best result from DPM-Solver++~\cite{lu2022dpm}, Uni\_PC~\cite{zhao2023unipc} and iPNDM~\cite{zhang2022DEISiPNDM} for DMN, GITS and LD3. (A further evaluation on varying solvers can be referred in the appendix.) For FLUX.1-dev and LTX-video, we set resolution dependent shifted timesteps (RDS)~\cite{flux2024} and globally optimized discretization(GOD, \ie, optimizing a single set of parameters) as our baseline.

\noindent\textbf{Evaluation metrics.} Our primary quantitative evaluation metric is the Fréchet Inception Distance (FID)~\cite{heusel2017gans} over 50k generated images. For text-to-image model (FLUX.1-dev), we calculate both FID and CLIP scores~\cite{radford2021learning}. These metrics are computed on a set of 10k generated images, using prompts randomly sampled from MS-COCO validation~\cite{lin2014microsoft} set, adhering to~\cite{wang2024taming}. Recognizing the potential limitations of FID for high-resolution text-to-image generation, we supplement our evaluation with CMMD~\cite{jayasumana2024cmmd} and provide qualitative analyses to ensure a comprehensive comparison. All quantitative metrics are averaged across three runs.

\noindent\textbf{Training settings}. We pre-generated fixed teacher datasets for distillation. For pixel-space and latent-space DPMs, this comprised 10,000 images and corresponding random generator states of noise produced by a 30-step iPNDM solver, serving as targets for student models across various NFEs. For FLUX.1-dev, 10,000 prompt-conditioned images were generated using a 10-step guided iPNDM solver. 
We also use gradient-checkpointing for reducing memory cost on Flux. For LTX-video, we use 5000 prompt-conditioned videos generated by a 7-step euler solver. Additional implementation details regarding efficiency can be referred to the appendix.


\begin{figure*}[t] 
    \centering 
    \vspace{-1.0cm}
    \includegraphics[width=1.0\textwidth]{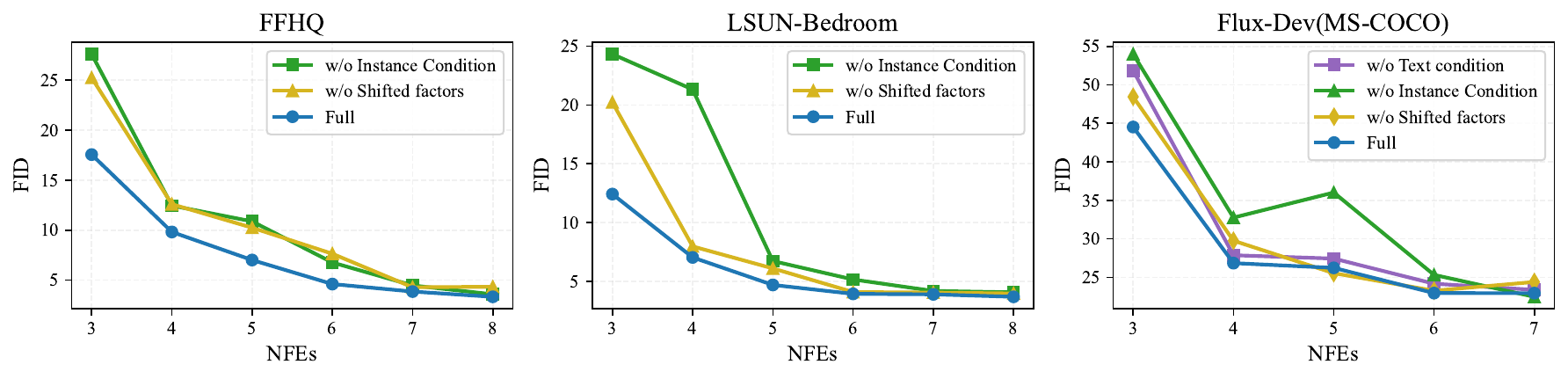}
    \vspace{-0.52cm}
    \caption{Ablation study on FFHQ, LSUN-Bedroom and FLUX.1-dev, instance condition is observed to be the most contributing factor, while the effect of shifted factors varies across pretrained models.}
    \vspace{-0.4cm}
    \label{fig:ablation}
\end{figure*}

\subsection{Main results}

\noindent\textbf{Pixel-Space DPMs.}
We first evaluate our instance-aware method against various discretization acceleration techniques, including those with discretization tuning and hand-crafted heuristics, on low-resolution pixel-space DPMs. As presented in~\Cref{tab:pixel_fid}, our approach, which conditions the discretization strategy on the initial prior sample, consistently outperforms the previous state-of-the-art methods. Specifically, compared to the strongest baseline, our method achieves average FID improvements of \(35.33\%\), \(31.50\%\), and \(15.62\%\) for NFE=3, 5, and 7, respectively, across four datasets. This trend indicates that the performance advantage of instance-aware discretization is more pronounced at lower NFEs, an observation consistent with our statistical analysis on the 2D synthetic examples in~\Cref{fig:quanttoy}.

%


\begin{wraptable}{r}{0.6\columnwidth}
    \centering
    \hspace{-0.5cm}
    \vspace{-0.3cm} 
    \renewcommand{\arraystretch}{1.2}
    \scriptsize
    \resizebox{\linewidth}{!}{
    \begin{tabular}{@{}lccc@{}}
    \toprule
    Method & NFE=3 & NFE=5 & NFE=7 \\
    \midrule
    best heu. & 41.99 & 6.38  & 4.39  \\
    DMN       & 28.11 & 6.15  & 5.16  \\
    AMED      & 58.21 & 13.20 & 7.00  \\
    GITS      & 44.78 & 17.29 & 9.59  \\
    LD3       & 14.62 & 5.93  & 4.31  \\
    \midrule
    \textbf{\Ours} & \textbf{12.44} & \textbf{4.99} & \textbf{3.81} \\
    \bottomrule
    \end{tabular}%
    }
    \hspace{-0.5cm}
    \caption{FID on latent space LSUN 256x256.}
    \label{tab:lsun_fid_main_wrap}
    \vspace{-0.4cm} 
\end{wraptable}
%

\noindent\textbf{Latent-Space DPMs}.
We further validate our instance-aware method on latent-space diffusion models (Stable Diffusion checkpoints on LSUN-Bedroom 256x256). The performance trends observed are consistent with those from pixel-space DPMs. For LSUN-Bedroom, we achieve an average FID improvement of \(14.12\%\) across NFE=3, 5, and 7 when compared to the strongest baseline results.


\begin{figure*}[t] 
    \centering 
    \includegraphics[width=1.0\textwidth]{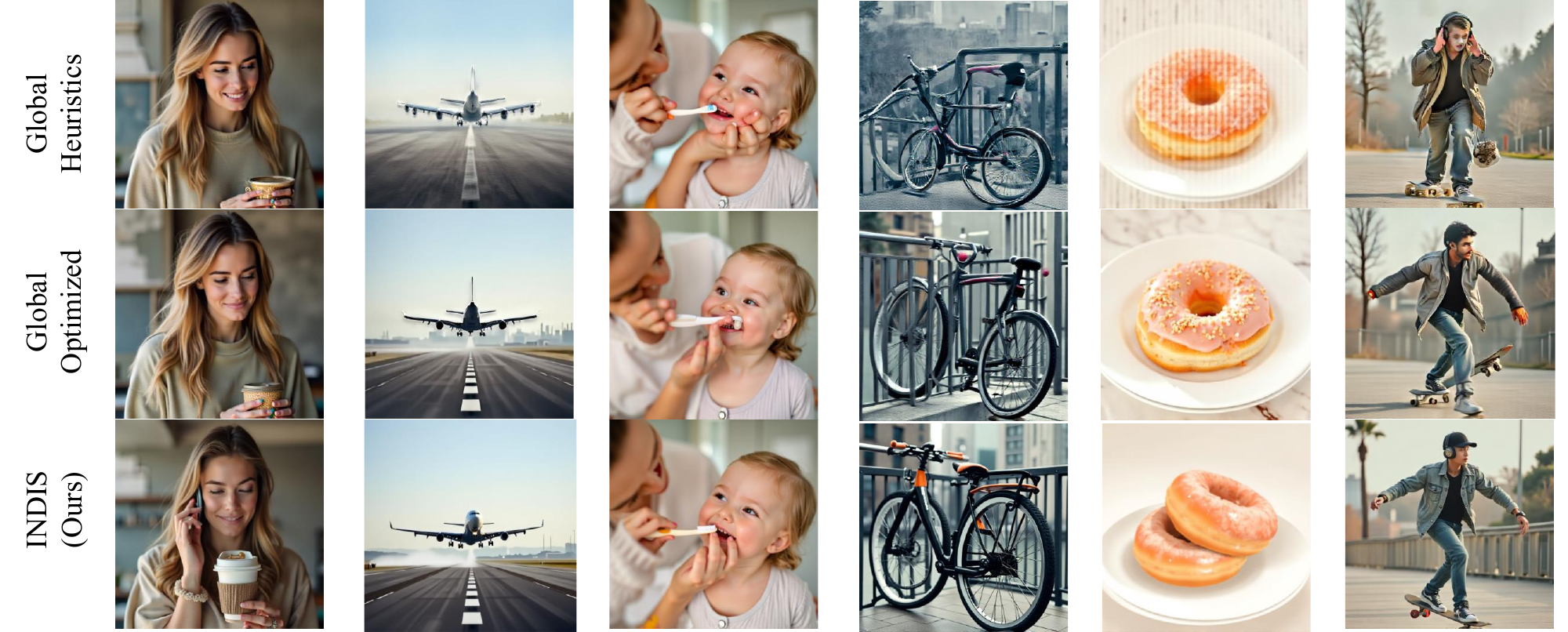}
    \caption{Qualitative Results on FlUX.1-dev (NFE=7) of our instance-level \(\Ours\) method, compared with Global Heuristics (RDS) and globally optimized discretization (GOD). Corresponding prompts could be referred in the appendix.}
    \vspace{-0.12cm}
    \label{fig:qualitative_flux}
\end{figure*}

\noindent\textbf{Comparison with solver distillation}. We also provide comparison with solver based distillation~\cite{wang2025adaptivestochasticcoefficientsaccelerating,zhu2025distilling} on pixel and latent domain.

\begin{table}[t]
\centering
\renewcommand{\arraystretch}{1.15}
\scriptsize
\resizebox{0.98\linewidth}{!}{%
\begin{tabular}{lccc|ccc}
\toprule
& \multicolumn{3}{c|}{\textbf{CIFAR10 (pixel-space)}} & \multicolumn{3}{c}{\textbf{LSUN 256$\times$256 (latent-space)}} \\
\textbf{Method} & \textbf{NFE=3} & \textbf{NFE=5} & \textbf{NFE=7} & \textbf{NFE=3} & \textbf{NFE=5} & \textbf{NFE=7} \\
\midrule
EPD    & 10.40 & 4.33 & 2.82 & 13.21 & 7.52 & 5.97 \\
AdaSDE & 12.62 & 4.18 & 2.88 & 18.03 & 6.96 & 5.16 \\
\textbf{\Ours} & \textbf{9.26} & \textbf{3.31} & \textbf{2.60} & \textbf{12.44} & \textbf{4.99} & \textbf{3.81} \\
\bottomrule
\end{tabular}%
}
\vspace{-0.18cm}
\caption{Comparison with solver distillation on FID.}
\vspace{-0.48cm}
\label{tab:solver_distill_cifar10_lsun}
\end{table}

\noindent\textbf{FLUX.1-dev and LTX-Video}. For FLUX.1-dev, we observe its inherent robustness to hyperparameter variations, leading to a diminishing impact of different discretization strategies as NFE increases. In few-NFE settings, our approach demonstrates its effectiveness compared to global counterparts. We also verify the effectiveness of our instance discretizations on LTX-Video~\cite{hacohen2024ltx} measured on VBench~\cite{huang2024vbench}. It's observed that given the latent prior of video, \(\Ours\) is able to capture instance-level benefits in discretizations, improving aesthetic, imaging quality, and subject consistency.

\vspace{-0.2cm}
\begin{table}[b] 
    \centering 
    \vspace{-0.6cm}
    \label{tab:flux_dev_main_comparison_wrap}
    \resizebox{0.45\textwidth}{!}{
    \renewcommand{\arraystretch}{1.2}
    \scriptsize
    \begin{tabular}{@{}clccc@{}}
    \toprule
    Metrics & Method & NFE=3 & NFE=5 & NFE=7 \\
    \midrule
    \multirow{3}{*}{\parbox[c]{1.5cm}{\centering FID ($\downarrow$)}} 
    & RDS & 64.50 & 30.12 & \textbf{22.58} \\
    & GOD     & 56.82 & 28.52 &  22.77 \\
    & \textbf{\Ours} & \textbf{ 44.35} & \textbf{24.89} & 22.70 \\
    \midrule
    \multirow{3}{*}{\parbox[c]{1.5cm}{\centering CLIP ($\uparrow$)}} 
    & RDS & 23.29 & 29.66 & 30.76 \\
    & GOD     & 24.41 & 29.70 & 30.80 \\
    & \textbf{\Ours} & \textbf{26.33} & \textbf{30.01} & \textbf{30.86} \\
    \midrule
    \multirow{3}{*}{\parbox[c]{1.5cm}{\centering CMMD ($\downarrow$)}} 
    & RDS & 1.75 & 0.86 & 0.89 \\
    & GOD     &1.72 & 0.79 & 0.75 \\
    & \textbf{\Ours} & \textbf{ 1.69} & \textbf{ 0.75} & \textbf{0.73} \\
    \bottomrule
    \end{tabular}%
    }
    \captionsetup{width=\linewidth} 
    \caption{Performance comparison on FLUX.1-dev (NFE=3, 5, 7).}
    \vspace{-0.35cm}
\end{table}

\begin{table}[htbp]
    \centering
    \renewcommand{\arraystretch}{1.1} 
    \resizebox{0.48\textwidth}{!}{
    \begin{tabular}{lccc}
    \toprule
    Method & Aes. Quality & Imaging Quality & Subject Consis. \\
    \midrule
    RDS & 0.579 & 0.597 & 0.963 \\
    GOD & 0.583 & 0.603 & 0.963 \\
    \textbf{INDIS} & \textbf{0.593} & \textbf{0.613} & \textbf{0.964} \\
    \bottomrule
    \end{tabular}
    }
    \caption{Performance comparison on LTX-Video (NFE=5).}
    \vspace{-1.0em}
    \label{tab:video}
\end{table}

\subsection{Ablations}

We ablate key framework components: \textit{w/o} instance-level conditioning, \textit{w/o} shift factors, and \textit{w/o} textual guidance (for FLUX.1-dev). As shown in \Cref{fig:ablation}, removing the instance-specific conditioning consistently causes the most severe performance drop. The effectiveness of shift and scale factors varies across base models. we attribute this to their differing noise schedules and inherent exposure bias severities (detailed in the appendix).

\begin{figure}[t]
    \centering
    \includegraphics[width=\linewidth]{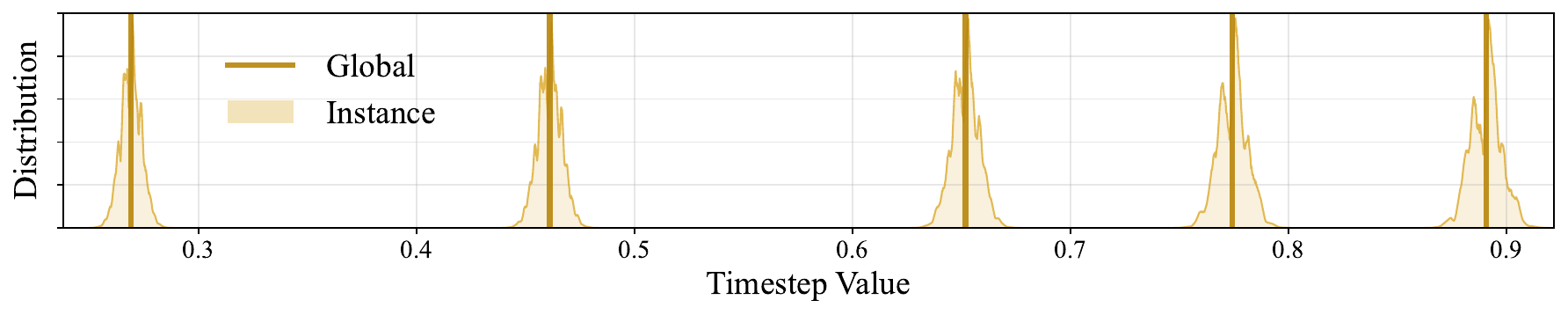}
    \caption{Visualization of instance-level discretizations on FLUX.}
    \label{fig:instance_visualization}
    \vspace{-0.45cm}
\end{figure}

\noindent\textbf{Qualitative results.} \Cref{fig:qualitative_flux} demonstrates that our instance-aware discretization (NFE=7) yields perceptibly stronger visual results on FLUX.1-dev, enhancing both detail and coherence over global heuristics. Additionally, \Cref{fig:instance_visualization} contrasts the distribution of our instance-level timesteps against global baselines on FLUX.1-dev.

\vspace{-0.3cm}
\section{Conclusion}\label{sec:conclusion}
\vspace{-0.2cm}

This work demonstrates the advantage of moving beyond globally fixed discretizations for diffusion ODE sampling. Motivated by synthetic experiments, we propose an instance-aware strategy that dynamically tailors timestep schedules to the initial noise and available guidance. Extensive evaluations across diverse diffusion and flow matching models for images and videos confirm that our approach consistently improves few-step sampling performance.

\noindent\textbf{Limitations and future work.} Relying on gradient checkpointing for scalability (\eg, on FLUX.1-dev) introduces computational overhead. Future work will explore integrating adjoint matching to optimize efficiency under specific solver parameterizations.






\section*{Acknowledgement}

This work was supported by the National Natural Science Foundation of China (No. 6250070674) and the Zhejiang Leading Innovative and Entrepreneur Team Introduction Program (2024R01007).

\bibliographystyle{unsrtnat}
\bibliography{ref}

\ifarxiv \clearpage \appendix \ifarxiv
\else
  \begin{document}
\fi

\ifarxiv
\else
  \title{\paperTitle}
  \author{\authorBlock}
  \maketitlesupplementary
\fi

\appendix

\tableofcontents

\section{Synthetic experiment details}
\label{sec:app_toyanalysis}

This section details the synthetic experiments that motivate our instance-specific framework. We begin by describing the setup of our 2D toy example and the different timestep optimization strategies under comparison (as illustrated in~\Cref{fig:toy_gaussian} and~\Cref{fig:quanttoy}). Then we provide additional qualitative comparison contributing to the motivation of our instance-level approach.

\subsection{Synthetic data and noise schedule configuration}
To better capture the distributional feature of high dimensional image data, we adhere to the synthetic data distribution used in~\cite{Karras2024autoguidance}. Based on this, we make the following modifications.

First, for better observation of  the transition trajectory from prior distribution to data distribution, we convert the variance exploding (VE) noise schedule \(\alpha_t = 1,\sigma_t = t\) to flow matching Optimal Transport(OT) noise schedule \(\alpha_t = 1-t,\sigma_t = t\). This ensures that the variance of prior and data distribution are on the same order of magnitude. Specifically, we make the following adaptation.
\begin{equation}
    t^{\text{OT}} = \frac{t^{\text{VE}}}{1+t^{\text{VE}}}, \quad \mathbf{x}_t^{\text{OT}} = \frac{1}{1+t^{\text{VE}}}\mathbf{x}_t^{\text{VE}}.
\label{eq:VEOT}
\end{equation}
The equivalence and transition between noise schedules are mathematically guaranteed, for readers interested, please refer to Proposition 1 in~\cite{lee2024improving} or Lemma 2 in~\cite{pokle2024training}. Subsequently, we convert the epsilon prediction \(\epsilon_{\theta}\)  to velocity prediction \(v_{\theta}\) (this can also be applied to data prediction \(\mathbf{x}_{\theta}\), we also refer the interested readers to~\cite{kingma2023understanding} for an in depth look):
\begin{align}
\begin{split}
v_{\theta}(\mathbf{x}^{\text{OT}}_t, t) &= \epsilon_{\theta}(\mathbf{x}^{\text{VE}}_t, t_{\text{VE}}) - \mathbf{x}_0\\
&= \epsilon_{\theta}(\mathbf{x}^{\text{VE}}_t, t_{\text{VE}}) - \frac{\mathbf{x}^{\text{OT}}_t - t^{\text{OT}}\cdot\epsilon_{\theta}(\mathbf{x}^{\text{VE}}_t, t_{\text{VE}})}{1-t^{\text{OT}}}\\
&= \frac{\epsilon_{\theta}(\mathbf{x}^{\text{VE}}_t, t_{\text{VE}}) - \mathbf{x}^{\text{OT}}_t}{1-t^{\text{OT}}}.
\end{split}
\end{align}

This transition provides the following benefits: the magnitude of the transition is preserved, qualitative comparison between trajectories becomes feasible, which provides insights for the design of our training dynamics.

\subsection{Comparison experiment setup}

Building upon the optimal transport noise schedule, we provide the detailed settings of (a), (b), (c), (d) in~\Cref{fig:toy_gaussian}. Specifically, we keep the training and sampling set for (b) (c) (d) identical. Given the NFE budget of \toyNstep,

\textbf{(a)} The timestep is uniformly discretized \(\{\tau_i = \frac{i}{N}(T-t_0) +t_0\}_{i=0}^{\toyNstep}\), serving as a baseline method. 

\textbf{(b)} The timestep is optimized through \Cref{eq:my_general_ts_optim} and shared during sampling. As,
\begin{equation}
    \label{eq:my_general_ts_optim_app}
    \arg\min_{\Studenttime} \mathbb{E}_{\mathbf{x}_T \sim \mathcal{N}(0,\sigma_T^2\mathbf{I})} \left[ \mathrm{d}( \ODE(\mathbf{x}_T,\Teachertime),\ODE(\mathbf{x}_T,\Studenttime))\right].
\end{equation}
\textbf{(c)} For each prior point, we conduct an instance-level optimization problem. During sampling, we assign each point the corresponding optimized timestep. Thus the optimization can be reframed as:
\begin{equation}
\label{eq:my_general_ts_optim_app2}
    \arg\min_{\{\Studenttime^{\mathbf{x}_T}\}} \mathbb{E}_{\mathbf{x}_T \sim \mathcal{N}(0,\sigma_T^2\mathbf{I})} \left[ \mathrm{d}( \ODE(\mathbf{x}_T,\Teachertime),\ODE(\mathbf{x}_T,\Studenttime^{\mathbf{x}_T}))\right].
\end{equation}
\textbf{(d)} The timestep is optimized through~\Cref{eq:my_general_ts_optim_instance}, the network design is simpler compared to high dimensional case, with 2 layers of FFN along with Relu activation, and apply sigmoid to normalize the output. Thus \(\Priornet(\cdot): \mathbb{R}^2 \to \mathbb{R}^{N}\). (Here \(2\) is the data dimension, \(N\) is the number of step.) 
\begin{equation}
\label{eq:my_general_ts_optim_instance_app}
    \arg\min_{\Priornet} \mathbb{E}_{\mathbf{x}_T \sim \mathcal{N}(0,\sigma_T^2\mathbf{I})} \left[ \mathrm{d}( \ODE(\mathbf{x}_T,\Teachertime),\ODE(\mathbf{x}_T,\Studenttime^{\Priornet}))\right].
\end{equation}
In~\Cref{fig:quanttoy}, uniform schedule represents (a), globally optimized represents (b) and instance-level condition represents (d).

\subsection{Results}

\noindent
\textbf{Quantitative results.} We conduct a comprehensive quantitative evaluation of three timestep scheduling strategies—uniform, globally optimized, and instance-level optimized—across different step budgets. As shown in Table~\ref{tab:schedule_comparison}, we assess each method using three metrics: KL divergence and Wasserstein distance to measure distribution-level fidelity, and mean squared error (MSE) to capture per-instance reconstruction accuracy. The results reveal that instance-level scheduling outperforms uniform and globally optimized method across all metrics and step counts, especially in low-NFE settings. Notably, instance-specific schedules achieve lower divergence and error with fewer steps, highlighting the benefits of dynamically adapting the timestep schedule to each sample.

\begin{table*}[t]
\centering
\caption{Comparison of different timestep scheduling strategies across MSE, KL divergence, and Wasserstein distance under varying step numbers.}
\label{tab:schedule_comparison}
\small
\setlength{\tabcolsep}{5pt}
\renewcommand{\arraystretch}{1.2}
\begin{tabular}{c | @{\hskip 6pt}ccc@{\hskip 6pt}|@{\hskip 6pt}ccc@{\hskip 6pt}|@{\hskip 6pt}ccc}
\toprule
\multirow{2}{*}{Steps} 
& \multicolumn{3}{c@{\hskip 6pt}|@{\hskip 6pt}}{\textbf{MSE ↓}} 
& \multicolumn{3}{c@{\hskip 6pt}|@{\hskip 6pt}}{\textbf{KL Divergence ↓}} 
& \multicolumn{3}{c}{\textbf{Wasserstein ↓}} \\
& Uniform & Global & \textbf{Instance} 
& Uniform & Global & \textbf{Instance}
& Uniform & Global & \textbf{Instance} \\
\midrule
3  & 0.0588 & 0.0245 & \cellcolor{gray!20}0.0158 & 5.1574 & 1.4452 & \cellcolor{gray!20}0.6091 & 0.2284 & 0.1466 & \cellcolor{gray!20}0.0925 \\
4  & 0.0213 & 0.0223 & \cellcolor{gray!20}0.0115 & 1.1362 & 1.3032 & \cellcolor{gray!20}0.6065 & 0.1361 & 0.1403 & \cellcolor{gray!20}0.0872 \\
5  & 0.0238 & 0.0122 & \cellcolor{gray!20}0.0093 & 1.2130 & 0.5819 & \cellcolor{gray!20}0.3463 & 0.1496 & 0.1039 & \cellcolor{gray!20}0.0771 \\
6  & 0.0147 & 0.0094 & \cellcolor{gray!20}0.0076 & 0.6775 & 0.3521 & \cellcolor{gray!20}0.2434 & 0.1139 & 0.0869 & \cellcolor{gray!20}0.0716 \\
7  & 0.0121 & 0.0072 & \cellcolor{gray!20}0.0066 & 0.4180 & 0.2246 & \cellcolor{gray!20}0.2194 & 0.1038 & 0.0787 & \cellcolor{gray!20}0.0722 \\
8  & 0.0098 & 0.0052 & \cellcolor{gray!20}0.0045 & 0.2871 & \cellcolor{gray!20}0.1143 & 0.1304 & 0.0953 & 0.0665 & \cellcolor{gray!20}0.0595 \\
9  & 0.0078 & 0.0046 & \cellcolor{gray!20}0.0033 & 0.2197 & 0.1257 & \cellcolor{gray!20}0.1026 & 0.0826 & 0.0622 & \cellcolor{gray!20}0.0510 \\
10 & 0.0065 & 0.0035 & \cellcolor{gray!20}0.0031 & 0.1630 & 0.0977 & \cellcolor{gray!20}0.0819 & 0.0791 & 0.0564 & \cellcolor{gray!20}0.0452 \\
\bottomrule
\end{tabular}
\end{table*}

\begin{figure*}[b] 
    \centering 
    \includegraphics[width=0.95\textwidth]{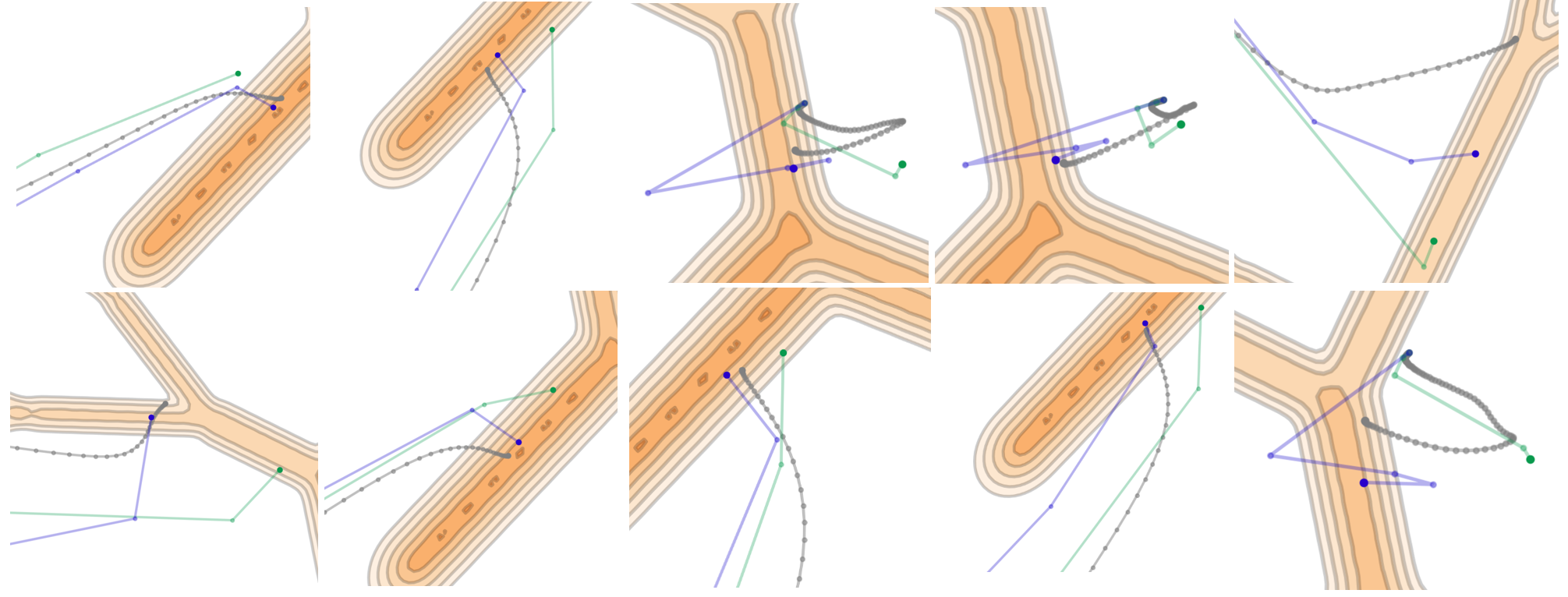}
    \caption{Qualitative comparison of sampling trajectories: Ground Truth (\textcolor{gray}{gray}, 100 NFE), Globally Optimized Timesteps (\textcolor{green!60!black}{green}), and Instance-Specific Timesteps (\textcolor{blue!60!purple}{purple}). Orange coutour represents the data manifold.}
    \label{fig:qualitativetoy}
\end{figure*}

\noindent
\textbf{Qualitative results.} As illustrated in~\Cref{fig:qualitativetoy}, we qualitatively compare the searched timesteps across different regions. This comparison reveals that while using a uniformly optimized timestep improves overall sample correctness, the instance-specific design provides greater flexibility. This allows for more tailored trajectories that better align with the ground truth sampled data points.

\noindent\textbf{Intermediary Supervision vs. Global error supervision} Our qualitative analysis presented in~\Cref{fig:qualitativetoy}, also offer insights regarding the choice of supervision signal for learning adaptive solver parameters. Different strategies exist in prior work: methods such as AMED~\cite{zhou2024fast} and Bespoke Solvers~\cite{shaul2024bespoke} utilize intermediary loss terms that compare states along the sampling trajectory. In contrast, approaches like LD3~\cite{tong2025LD3} and Bespoke Non-stationary Solvers compute the distance metric only at the final endpoint $\mathbf{x}_0$, effectively supervising based on the global truncation error. The strong performance achieved with global error supervision supported by arguments in LD3~\cite{tong2025LD3} and theoretical validation in AYS~\cite{sabour2024align}, may be attributed to its robustness. Specifically, when there is a substantial NFE gap between student and teacher solvers, their intermediate trajectories can diverge significantly, potentially making intermediary supervision signals less reliable or even misleading. Global error supervision, by focusing only on the final outcome, may provide the optimization process with a larger effective search space and more flexibility in determining the parameterization for the entire path.

\section{Implementation and architectural specifics}\label{sec:implementation_details}

\subsection{Prior conditioning network architecture}\label{subsec:architecture}

Here we present a detailed description of our prior conditioning network.

Our instance-aware parameter prediction network takes a processed representation of the initial noise $\mathbf{x}_T$, combined with an embedding of any available conditional guidance $\mathbf{c}$, as its input. The initial noise $\mathbf{x}_T \sim \mathcal{N}(0, \sigma_T^2 \mathbf{I})$ is first normalized to ensure unit variance, agnostic to noise schedule. We then apply Singular Value Decomposition (SVD), $\text{SVD}(\mathbf{x}_T) \rightarrow U\Sigma V^T$ for feature rearrangement. The resulting components—the singular vectors $U, V^T$ and singular values $\Sigma$—are individually processed through FFNs with ReLU activation, and their outputs are subsequently concatenated to form the noise representation, $\text{Rep}(\mathbf{x}_T)$.

\begin{figure*}[ht] 
    \centering 
    \includegraphics[width=0.88\textwidth]{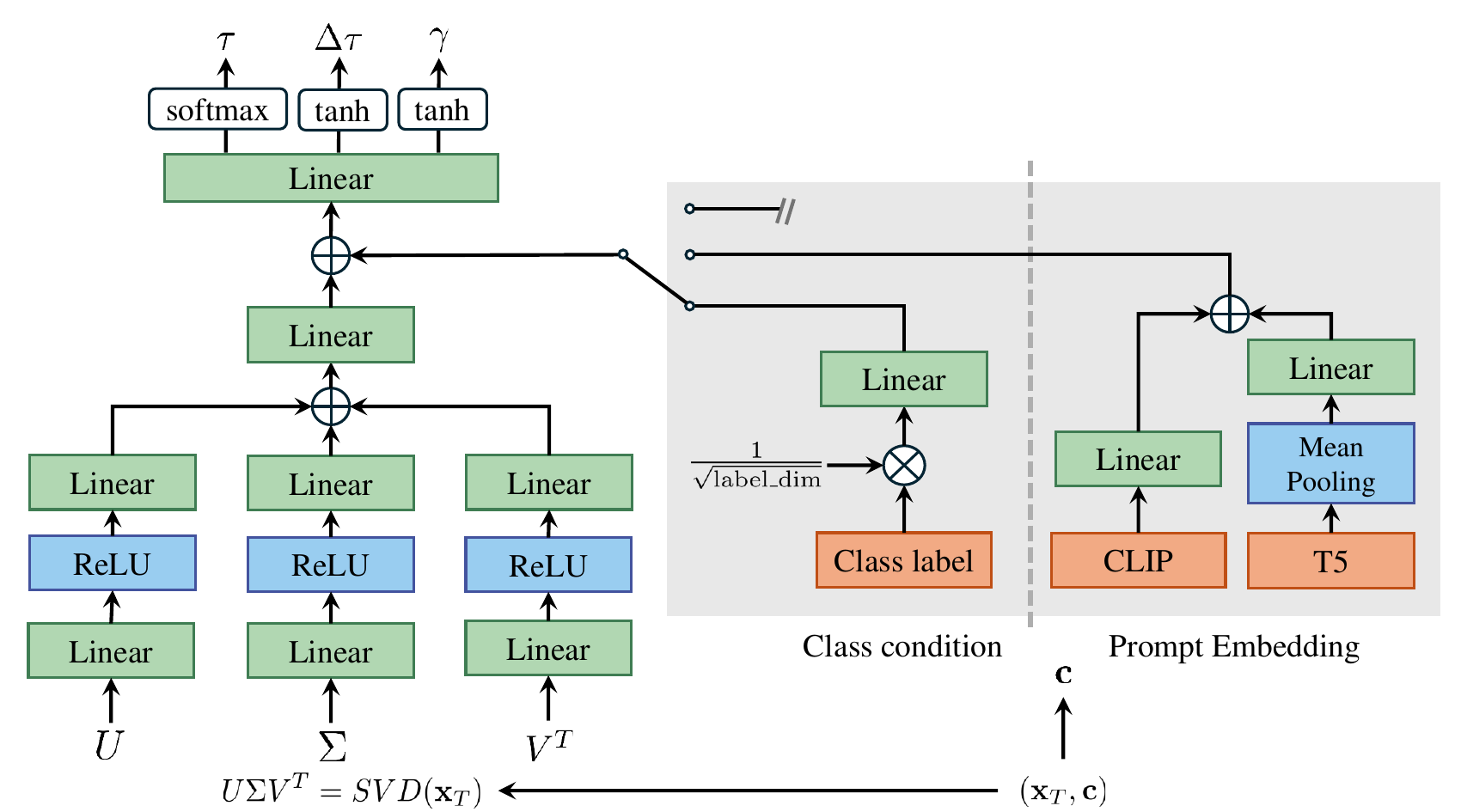}
    \caption{Architectural design of the proposed lightweight prior conditioning network. When conditional information is available, class indices are first scaled by a factor of $\frac{1}{\sqrt{\text{label\_dim}}}$ and then processed through a linear layer. For prompt embeddings (FLUX.1-dev), T5 embeddings undergo mean pooling to reduce dimensionality before being concatenated with CLIP embeddings.}
    \label{fig:network_supp}
\end{figure*}

Conditional guidance $\mathbf{c}$ is transformed into a suitable embedding, $Emb(\mathbf{c})$, before being combined with $\text{Rep}(\mathbf{x}_T)$. For class labels, the one-hot encoded vector is scaled by a factor of $1/\sqrt{\text{dim}_{\text{label}}}$ to ensure unit variance, following recommendations in~\cite{karras2024edm2}. For text-based conditioning, exemplified by architectures such as FLUX.1-dev DiT which may utilize dual text embeddings, each text embedding is passed through Linear layers. These processed text features are then combined (e.g., via concatenation or summation) to form the unified $Emb(\mathbf{c})$. Finally, the representations $\text{Rep}(\mathbf{x}_T)$ and $Emb(\mathbf{c})$ are concatenated to serve as the complete input to our parameter prediction network. For video latent structures with \([f,C,H,W]\), directly handling the latent requires heavy computational overhead compared to images, thus we first pool the video latent on the frame dimension \(f\) to ensure computational efficiency. 
The overall architectural design illustrated in~\Cref{fig:network_supp}.

\subsection{Mitigating exposure bias via shift factors}
\label{subsec:shiftfactor}

A common challenge in diffusion models is \textit{exposure bias}. During training, the network ($\epsilon_\theta$) is exposed to noisy states $\mathbf{x}_t$ derived directly from clean data $\mathbf{x}_0$ (i.e., $\mathbf{x}_t = \alpha_t \mathbf{x}_0 + \sigma_t \epsilon$). During sampling, the network processes states generated iteratively from an initial noise $\mathbf{x}_T$, leading to states that can drift from the distribution seen in training. This discrepancy degrades performance, especially with few sampling steps. This mismatch problem is noted in prior literature~\cite{ning2024elucidating,lin2024common,li2024alleviating}. Our learnable shift and scale factors are designed to mitigate such effects.
We detail three noise schedules and their Signal-to-Noise Ratios (SNRs) at the starting point of sampling process, $t=T$.

\noindent\textbf{EDM-VE.} The Elucidating DPM (EDM) framework utilizes a Variance Exploding (VE) schedule where $\alpha_t = 1$ and $\sigma_t = t$. For the maximum time $T_{max} = 80.0$ (where sampling begins), the SNR (defined as $\alpha_{T}^2 / \sigma_{T}^2$) is $1/80^2 = 1.5625 \times 10^{-4}$.

\noindent\textbf{Stable Diffusion-VP.} This Variance Preserving (VP) schedule is an adaptation of the DDPM linear schedule. With $\beta(t)$ representing the noise variance schedule in continuous time, the schedule parameters are $\alpha_t = \exp(-\frac{1}{2}\int_0^t \beta(s) ds)$ and $\sigma_t = \sqrt{1-\alpha_t^2}$. Using the specific continuous $\beta(t) = (\sqrt{0.00085}\cdot(1-t) + \sqrt{0.012}\cdot t)^2$ (where $t$ is normalized to $[0,1]$, and sampling starts at $t=1$), the resulting SNR is $4.7\times10^{-3}$.

\noindent\textbf{Flow Matching-OT.} The noise schedule is $\alpha_t =1-t,\sigma_t = t$. While the full implementation(training) of Flux~\cite{flux2024} is not available, the sampling implementation suggests a starting timestamps of \(T = 1.0\).

To alleviate exposure bias problem in few step sampling, we make the following adaptation: given the function evaluation at current state \(\epsilon_{\theta}(\mathbf{x}_{\tau_n},\tau_n)\), we introduce the shift factors in the following form.
\begin{align}
\begin{split}
    \hat{\epsilon}_{\theta}(\mathbf{x}_n,\tau_n,\Delta \tau_n, \gamma_n) := \gamma_n\cdot\epsilon_{\theta}(\mathbf{x}_n,\tau_n+\Delta \tau_n), \\ \Studenttime^{\Priornet} = \{\tau_n,\Delta\tau_n, \gamma_n\}_{n=1}^N = \phi(\mathbf{x}_T, \mathbf{c}).
\end{split}
\end{align}
\noindent\textbf{Observations.} \Cref{fig:ablation} in the paper demonstrates that the learnable shift and scale factors yield a substantially greater impact when applied to models utilizing the EDM-VE schedule (\eg, on FFHQ dataset) and DDPM-VP in LDM (\eg, on LSUN-bedroom), compared to FLUX.1-dev. In FLUX.1-dev, 
as NFE increases, the effect of shift factors becomes more negligible and even negative compared to the other two pretrained model.
We attribute this disparity primarily to the more pronounced exposure bias inherent in EDM-VE and VP (Stable Diffusion) schedule, which provides a larger scope for improvement through our shift factor design.

\subsection{Training configuration and sampling efficiency} 
\label{subsec:training_dynamics}

The optimization of a set of hyperparameters is known to be challenging, often exhibiting instability and necessitating meticulous design to hit optimal configurations~\cite{tong2025LD3}. We alleviate this by designing an instance level network, 
to further improve and stabilize the training procedure of our prior conditioning network, we adhere to the following settings.

\noindent\textbf{Training configuration}. We pre-generate a fixed teacher dataset of noise data pairs, similar to LD3~\cite{tong2025LD3}. Empirically, a larger pre-generated dataset improves our model's performance. We test among Dpm\_Solver, Uni\_PC and iPNDM and select the best based on FID as our teacher trajectory. We empirically find that iPNDM gives the most promising teacher data. To save memory, we save the random generator state instead of raw gaussian noise. Building upon efficiency consideration, we adopt the Analytical First Step (AFS)~\cite{dockhorn2022genie}, which analytically approximates the initial update without invoking the denoising network, reducing the total NFE by one. Additionally, to stabilize optimization under limited NFE budgets, we employ Exponential Moving Average (EMA) on \(\Priornet\), comparing configurations with and without EMA (20\%) and selecting the best-performing variant based on validation results. We present the hyperparameter setting as in~\Cref{tab:hyperparameter_settings}.

\begin{table}[!b]
    \centering
    \captionsetup{font=footnotesize}
    \renewcommand{\arraystretch}{1.12}
    \scriptsize

    \resizebox{\columnwidth}{!}{%
    \begin{tabular}{lccccccc}
        \toprule
        Hyperparameter & CIFAR10 & FFHQ & AFHQv2 & ImageNet64 & LSUN & FLUX.1-dev & LTX-Video \\
        \midrule
        Learning Rate     & 0.05 & 0.05 & 0.05 & 0.05 & 0.05 & 0.001 & 0.001 \\
        Batch Size        & 16   & 16   & 16    & 8    & 4    & 4     & 4 \\
        GPUs (A100)       & 1    & 1    & 1     & 1    & 4    & 4     & 4 \\
        Nimgs/videos      & 10k  & 10k  & 10k   & 10k  & 10k  & 10k   & 5k \\
        $b_{\gamma}$      & 0.05 & 0.05 & 0.05  & 0.05 & 0.05 & 0.01  & 0.01 \\
        $b_{\Delta\tau}$  & 0.05 & 0.05 & 0.05  & 0.05 & 0.05 & 0.01  & 0.01 \\
        \bottomrule
    \end{tabular}%
    }

    \vspace{-0.4em}
    \caption{Hyperparameter settings.}
    \label{tab:hyperparameter_settings}
    \vspace{-0.8em}
\end{table}

\noindent\textbf{Efficiency analysis.}
Table~\ref{tab:efficiency_analysis} presents an efficiency analysis of our instance-aware parameter network. Training times reported are evaluated on NVIDIA A100 GPUs. The Sampling Overhead (\%) column quantifies the ratio of our prior network's inference time to the total sampling time; this specific overhead is evaluated for an NFE=5 setting. Parameter Overhead (\%) is calculated as the ratio of our prior network's parameters to those of the base diffusion model. The sampling overhead for FLUX.1-dev is \(2.3\%\), and CIFAR10 is \(2.5\%\) and ImageNet64 is \(1.9\%\).

\begin{table}[htbp] 
    \centering 
    \resizebox{0.98\columnwidth}{!}{
    \renewcommand{\arraystretch}{1.2}
    \begin{tabular}{lccc}
        \toprule
        Models & Training Time & Sampling Overhead (\%) & Parameter Overhead (\%) \\ 
        \midrule
        CIFAR10 (\(32\times 32\)) & 8 min & 2.5\% & 0.43\% \\
        ImageNet64 (\(64\times 64\)) & 40 min & 1.9\% & 0.32\% \\
        FLUX.1-dev (\(512\times 512\)) & 8 h$^{\dag}$ & 2.3\% & 0.21\% \\ 
        \bottomrule
    \end{tabular}
    }
    \vspace{-0.15cm}
    \caption{Efficiency Analysis. ${\dag}$: for FLUX.1-dev, we train \(\Priornet\) for 2 hours on 4 A100 GPUs with a batch size of 4. } %
    \label{tab:efficiency_analysis} 
\end{table}

\section{Additional experimental results}\label{sec:app_experiments}

Here we first provide ablation experiments regarding the design components of our instance-specific paradigm. Then we provide the full experimental results across various NFE (3-8) settings, and additional qualitative results.

\subsection{Ablations on design components.}

\begin{figure*}[b]
    \centering
    \includegraphics[width=1.9\columnwidth]{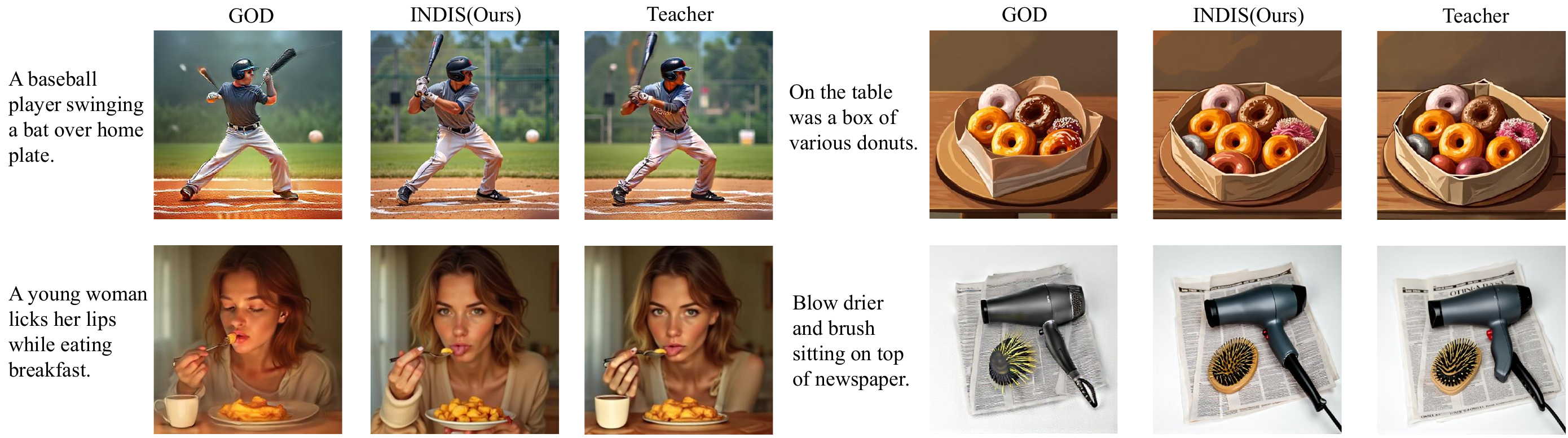}
    \caption{Qualitative comparison against teacher and global baseline.}
    \label{fig:indis_teacher}
\end{figure*}

\noindent\textbf{Ablation instance level framework.} 
We first ablate the design components in our INDIS framework: Singular value decomposition of the noise, discretization shifted factors and the instance-network itself. As illustrated in~\Cref{tab:component_ablation}. It's observed that both shifted factors and instance-level design are crucial for the final results, with instance-level discretization influencing the majority of the performance. 
\begin{table}[htbp]
    \centering
    \resizebox{0.48\textwidth}{!}{%
    \renewcommand{\arraystretch}{1.1} 
    \begin{tabular}{lcccccc}
    \toprule
    & \multicolumn{6}{c}{NFE} \\
    \cmidrule(lr){2-7}
    Ablation (on CIFAR-10) & 3 & 4 & 5 & 6 & 7 & 8 \\
    \midrule
    w/o instance level & 12.05 & 14.44 & 5.36 & 3.83 & 3.84 & 3.20 \\
    w/o shifted factors & 11.12 & 8.35 & 4.64 & 3.30 & 2.92 & 3.24 \\
    w/o SVD & 9.60 & 4.31 & \textbf{3.27} & \textbf{2.65} & 2.77 & 2.62 \\
    \textbf{full} & \textbf{9.26} & \textbf{4.14} & 3.31 & 2.81 & \textbf{2.60} & \textbf{2.34} \\
    \bottomrule
    \end{tabular}
    }
    \caption{Ablation study on the components of our method. Metrics are FID $\downarrow$.}
    \label{tab:component_ablation}
\end{table}

\noindent\textbf{Ablation on solver choices.} 
The improvements of instance-aware approach is agnostic to solver choices (as illustrated in~\Cref{tab:solver_ablation}, all solver choices reach comparable few-step generation results.), while the selection of iPNDM is that it serves as a good foundation for quality improvement.

\begin{table}[htbp]
    \centering
    \resizebox{0.48\textwidth}{!}{%
    \renewcommand{\arraystretch}{1.1} 
    \begin{tabular}{lcccccc}
    \toprule
    & \multicolumn{6}{c}{NFE} \\
    \cmidrule(lr){2-7}
    Solver choices & 3 & 4 & 5 & 6 & 7 & 8 \\
    \midrule
    Uni\_PC & 68.25 & 43.92 & 24.01 & 13.12 & 6.63 & 4.41 \\
    Uni\_PC (+INDIS) & 10.88 & 5.88 & 3.84 & 3.57 & \textbf{2.59} & 2.48 \\
    \midrule
    DPM-solver++ & 68.43 & 46.59 & 24.99 & 12.16 & 6.88 & 4.62 \\
    DPM-solver++ (+INDIS) & 9.92 & 6.52 & 4.62 & 3.55 & 3.17 & 2.92 \\
    \midrule
    iPNDM & 57.39 & 29.78 & 17.35 & 9.95 & 7.61 & 5.41 \\
    iPNDM (+INDIS) & \textbf{9.26} & \textbf{4.14} & \textbf{3.31} & \textbf{2.81} & 2.60 & \textbf{2.34} \\
    \bottomrule
    \end{tabular}
    }
    \caption{Ablation study on different solver choices, with and without our method (INDIS), on CIFAR-10.}
    \label{tab:solver_ablation}
\end{table}

\noindent\textbf{Ablation on teacher steps.} We first ablate our teacher-steps design choices, ranging from 10-30. The results (\Cref{tab:ablation_teacherstep}) demonstrate a direct correlation between teacher precision and student performance. We select the 30-step iPNDM solver as the default setting in our pixel space diffusion models and latent space LSUN-Bedroom.

\begin{table}[htbp]
    \centering
    \resizebox{0.48\textwidth}{!}{%
    \renewcommand{\arraystretch}{1.1} 
    \begin{tabular}{lcccccc}
    \toprule
    
    Teacher NFEs.& 3 & 4 & 5 & 6 & 7 & 8 \\
    \midrule
    10 & 11.57 & 8.34 & 5.52 & 4.01 & 4.45 & 3.38 \\
    20 & 9.73 & 6.59 & 3.48 & 3.11 & 2.81 & 2.55 \\
    30 & \textbf{9.26} & \textbf{4.14} & \textbf{3.31} & \textbf{2.81} & \textbf{2.60} & \textbf{2.34} \\
    \bottomrule
    \end{tabular}
    }
    \caption{Ablations on teacher solver steps.}
    \label{tab:ablation_teacherstep}
\end{table}

\subsection{Results on pixel space DPMs}

Here we present the full results of pixel space DPMs across NFEs, as illustrated in~\Cref{tab:cifar10_fid_comparison,tab:ffhq_fid_comparison,tab:afhqv2_fid_comparison,tab:imagenet64_fid_comparison}.

\begin{table*}[t]
\centering
\captionsetup{font=footnotesize}

\begin{minipage}[t]{0.49\textwidth}
\centering
\renewcommand{\arraystretch}{1.2}
\scriptsize
\resizebox{\linewidth}{!}{%
\begin{tabular}{lcccccc}
\toprule
Method &NFE=3 &NFE=4 &NFE=5 &NFE=6 &NFE=7 &NFE=8 \\
\midrule
best heu. &57.39 &29.78 &17.35 &9.95 &7.61 &5.41 \\
DMN &77.69 &26.35 &12.93 &8.09 &5.42 &5.90 \\
AMED &18.49 &17.18 &7.59 &7.04 &4.36 &5.56 \\
GITS &25.98 &10.11 &6.77 &4.29 &3.43 &2.70 \\
LD3 &16.52 &9.31 &5.32 &3.35 &3.37 &2.65 \\
\midrule
\textbf{\Ours} &\textbf{9.26} &\textbf{4.14} &\textbf{3.31} &\textbf{2.81} &\textbf{2.60} &\textbf{2.34} \\
\bottomrule
\end{tabular}}
\captionof{table}{FID results on CIFAR10.}
\label{tab:cifar10_fid_comparison}
\end{minipage}
\hfill
\begin{minipage}[t]{0.49\textwidth}
\centering
\renewcommand{\arraystretch}{1.2}
\scriptsize
\resizebox{\linewidth}{!}{%
\begin{tabular}{lcccccc}
\toprule
Method &NFE=3 &NFE=4 &NFE=5 &NFE=6 &NFE=7 &NFE=8 \\
\midrule
best heu. &72.29 &29.35 &17.52 &11.44 &8.76 &6.86 \\
DMN &178.09 &31.30 &20.93 &12.12 &10.17 &11.00 \\
AMED &26.87 &26.89 &12.49 &9.97 &6.64 &7.86 \\
GITS &26.41 &13.59 &8.85 &6.39 &5.36 &4.91 \\
LD3 &23.86 &14.15 &8.56 &5.97 &4.69 &3.97 \\
\midrule
\textbf{\Ours} &\textbf{17.72} &\textbf{8.92} &\textbf{6.91} &\textbf{4.72} &\textbf{3.90} &\textbf{3.31} \\
\bottomrule
\end{tabular}}
\captionof{table}{FID results on FFHQ.}
\label{tab:ffhq_fid_comparison}
\end{minipage}

\vspace{0.35cm}

\begin{minipage}[t]{0.49\textwidth}
\centering
\renewcommand{\arraystretch}{1.2}
\scriptsize
\resizebox{\linewidth}{!}{%
\begin{tabular}{lcccccc}
\toprule
Method &NFE=3 &NFE=4 &NFE=5 &NFE=6 &NFE=7 &NFE=8 \\
\midrule
best heu. &40.24 &15.35 &9.01 &6.26 &4.73 &3.83 \\
DMN &178.76 &33.15 &26.11 &16.01 &13.03 &10.12 \\
AMED &31.82 &18.99 &7.34 &8.19 &4.39 &5.55 \\
GITS &24.17 &12.20 &8.72 &6.10 &5.48 &4.03 \\
LD3 &17.94 &9.33 &6.09 &3.63 &2.77 &2.63 \\
\midrule
\textbf{\Ours} &\textbf{10.15} &\textbf{4.80} &\textbf{3.48} &\textbf{2.77} &\textbf{2.37} &\textbf{2.35} \\
\bottomrule
\end{tabular}}
\captionof{table}{FID results on AFHQv2.}
\label{tab:afhqv2_fid_comparison}
\end{minipage}
\hfill
\begin{minipage}[t]{0.49\textwidth}
\centering
\renewcommand{\arraystretch}{1.2}
\scriptsize
\resizebox{\linewidth}{!}{%
\begin{tabular}{lcccccc}
\toprule
Method &NFE=3 &NFE=4 &NFE=5 &NFE=6 &NFE=7 &NFE=8 \\
\midrule
best heu. &44.93 &21.32 &15.53 &10.27 &8.64 &6.60 \\
DMN &33.72 &15.24 &10.47 &6.74 &5.39 &4.98 \\
AMED &28.06 &32.69 &10.74 &10.63 &6.66 &7.71 \\
GITS &26.41 &16.41 &9.85 &8.39 &6.44 &5.64 \\
LD3 &27.82 &17.03 &11.55 &7.53 &5.63 &5.40 \\
\midrule
\textbf{\Ours} &\textbf{18.96} &\textbf{10.95} &\textbf{7.28} &\textbf{5.82} &\textbf{4.94} &\textbf{4.49} \\
\bottomrule
\end{tabular}}
\captionof{table}{FID results on ImageNet64.}
\label{tab:imagenet64_fid_comparison}
\end{minipage}

\end{table*}

\begin{table*}[t]
\centering
\renewcommand{\arraystretch}{1.15}
\scriptsize
\resizebox{\textwidth}{!}{%
\begin{tabular}{lccc|ccc|ccc|ccc}
\toprule
& \multicolumn{3}{c|}{\textbf{CIFAR10 32$\times$32}} 
& \multicolumn{3}{c|}{\textbf{FFHQ 64$\times$64}} 
& \multicolumn{3}{c|}{\textbf{ImageNet64 (class-cond.)}} 
& \multicolumn{3}{c}{\textbf{LSUN 256$\times$256 (latent)}} \\
\textbf{Method} 
& \textbf{NFE=3} & \textbf{NFE=5} & \textbf{NFE=7}
& \textbf{NFE=3} & \textbf{NFE=5} & \textbf{NFE=7}
& \textbf{NFE=3} & \textbf{NFE=5} & \textbf{NFE=7}
& \textbf{NFE=3} & \textbf{NFE=5} & \textbf{NFE=7} \\
\midrule
EPD
& 10.40 & 4.33 & 2.82
& 21.74 & 7.84 & 4.81
& \textbf{18.28} & \textbf{6.35} & 5.26
& 13.21 & 7.52 & 5.97 \\
AdaSDE
& 12.62 & 4.18 & 2.88
& 23.80 & 8.05 & 5.11
& 18.51 & 6.90 & 5.26
& 18.03 & 6.96 & 5.16 \\
\textbf{\Ours}
& \textbf{9.26} & \textbf{3.31} & \textbf{2.60}
& \textbf{17.72} & \textbf{6.91} & \textbf{3.90}
& 18.96 & 7.28 & \textbf{4.94}
& \textbf{12.44} & \textbf{4.99} & \textbf{3.81} \\
\bottomrule
\end{tabular}%
}
\vspace{0.08cm}
\caption{Comparison with solver distillation. FID at NFE=3,5,7 on CIFAR10, FFHQ, class-conditional ImageNet64, and LSUN 256$\times$256 (latent-space).}
\label{tab:appendix_solver_distill}
\end{table*}

\subsection{Results on latent space DPMs}

Here we present the full results of latent space DPMs and flow matching models across NFEs, as illustrated in~\Cref{tab:lsun_fid_comparison,tab:flux_dev_comparison}

\begin{table}[htbp] 
\centering
\resizebox{0.5\textwidth}{!}{%
\renewcommand{\arraystretch}{1.2}%
\scriptsize
\begin{tabular}{lcccccc} 
\toprule
Method &NFE=3 &NFE=4 &NFE=5 &NFE=6 &NFE=7 &NFE=8 \\
\midrule
best heu. &41.99 &11.93 &6.38 &5.08 &4.39 &4.88 \\
DMN &28.11 &11.82 &6.15 &4.71 &5.16 &4.55 \\
AMED &58.21 &15.67 &13.20 &8.92 &7.00 &4.19 \\
GITS &44.78 &21.67 &17.29 &11.52 &9.59 &8.82 \\
LD3 &14.62 &8.48 &5.93 &4.52 &4.31 &4.22 \\ 
\midrule
\textbf{\Ours} &\textbf{12.44} &\textbf{6.55} &\textbf{4.99} &\textbf{3.84} &\textbf{3.81} &\textbf{3.66} \\
\bottomrule
\end{tabular}%
}
\vspace{0.15cm}
\captionsetup{width=0.8\textwidth}
\caption{FID results on LSUN-Bedroom.}
\label{tab:lsun_fid_comparison}
\end{table}

\begin{table}[htbp]
\centering

\resizebox{0.5\textwidth}{!}{%
\renewcommand{\arraystretch}{1.2}
\scriptsize
\begin{tabular}{clcccccc}\toprule
Metrics & Method & NFE=3 & NFE=4 & NFE=5 & NFE=6 & NFE=7 \\\midrule
\multirow{3}{*}{\parbox[c]{1.2cm}{\centering FID($\downarrow$)}}
& RDS & 64.50 & 30.36 & 30.12 & 23.16 & \textbf{22.58} \\
& GOD & 56.82 & 29.33 & 28.52 & 23.32 & 22.77 \\
& \Ours & \textbf{44.35} & \textbf{26.07} & \textbf{24.89} & \textbf{22.93} & 22.70 \\
\midrule
\multirow{3}{*}{\parbox[c]{1.2cm}{\centering CLIP($\uparrow$)}}
& RDS & 23.29 & 27.77 & 29.66 & 30.42 & 30.76 \\
& GOD & 24.41 & 28.24 & 29.70 & 30.55 & 30.80 \\
& \Ours & \textbf{26.33} & \textbf{28.70} & \textbf{30.01} & \textbf{30.67} & \textbf{30.86} \\
\midrule
\multirow{3}{*}{\parbox[c]{1.2cm}{\centering CMMD($\downarrow$)}}
& RDS & 1.75 & 1.10 & 0.86 & 0.82 & 0.89 \\
& GOD & 1.72 & 1.01 & 0.79 & 0.77 & 0.75 \\
& \Ours & \textbf{1.69} & \textbf{0.98} & \textbf{0.75} & \textbf{0.73} & \textbf{0.73} \\
\bottomrule
\end{tabular}%
}
\vspace{0.15cm}
\caption{Full comparison on FLUX.1-dev(MS-COCO). For each column, we bold the best performing method. We found AFS to be negative on this large scale pretrained model, thus reporting the result w/o AFS.}
\label{tab:flux_dev_comparison}
\end{table}


\subsection{Results on Video Domains}

Besides VBench~\cite{huang2024vbench}, we also extend our instance-aware discretization strategy to VMBench~\cite{ling2025vmbench}, to further test the robustness of our method as illustrated in~\Cref{tab:video_appendix}.

\begin{table}[t]
\centering
\renewcommand{\arraystretch}{1.18}
\scriptsize
\resizebox{1.0\columnwidth}{!}{%
\begin{tabular}{llcccccc}
\toprule
\textbf{Prompt set} & \textbf{Method}
& \textbf{Aes.} & \textbf{Img.} & \textbf{Subj.}
& \textbf{PAS} & \textbf{TCS} & \textbf{CAS} \\
\midrule
\multirow{3}{*}{\textbf{VBench}}
& RDS & 0.579 & 0.597 & 0.963 & 2.162 & 95.643 & 49.615 \\
& GOD & 0.583 & 0.603 & 0.963 & 3.155 & 96.672 & 49.776 \\
& \textbf{INDIS} & \textbf{0.593} & \textbf{0.613} & \textbf{0.964} & \textbf{3.368} & \textbf{98.337} & \textbf{49.852} \\
\midrule
\multirow{3}{*}{\textbf{VMBench}}
& RDS & 0.552 & 0.601 & 0.955 & 2.956 & 96.053 & 48.981 \\
& GOD & 0.561 & 0.607 & 0.956 & 3.220 & 96.892 & 49.693 \\
& \textbf{INDIS} & \textbf{0.583} & \textbf{0.611} & \textbf{0.961} & \textbf{3.347} & \textbf{97.187} & \textbf{50.138} \\
\bottomrule
\end{tabular}%
}
\vspace{-0.6em}
\captionsetup{font=footnotesize}
\caption{Performance on VBench and VMBench prompt subsets.}
\vspace{-1.2em}
\label{tab:video_appendix}
\end{table}

\subsection{Comparison with Solver Distillation}

Here we provide additional comparison with solver distillation approaches, including the efficient parallel gradients method EPD~\cite{zhu2025distilling}, and an SDE learning based variant AdaSDE~\cite{wang2025adaptivestochasticcoefficientsaccelerating}, the result is illustrated in~\Cref{tab:appendix_solver_distill}.

\section{Qualitative Comparison}
\label{sec:qualitative_comparison}

\begin{figure*}[b]
    \centering
    \includegraphics[width=0.85\linewidth]{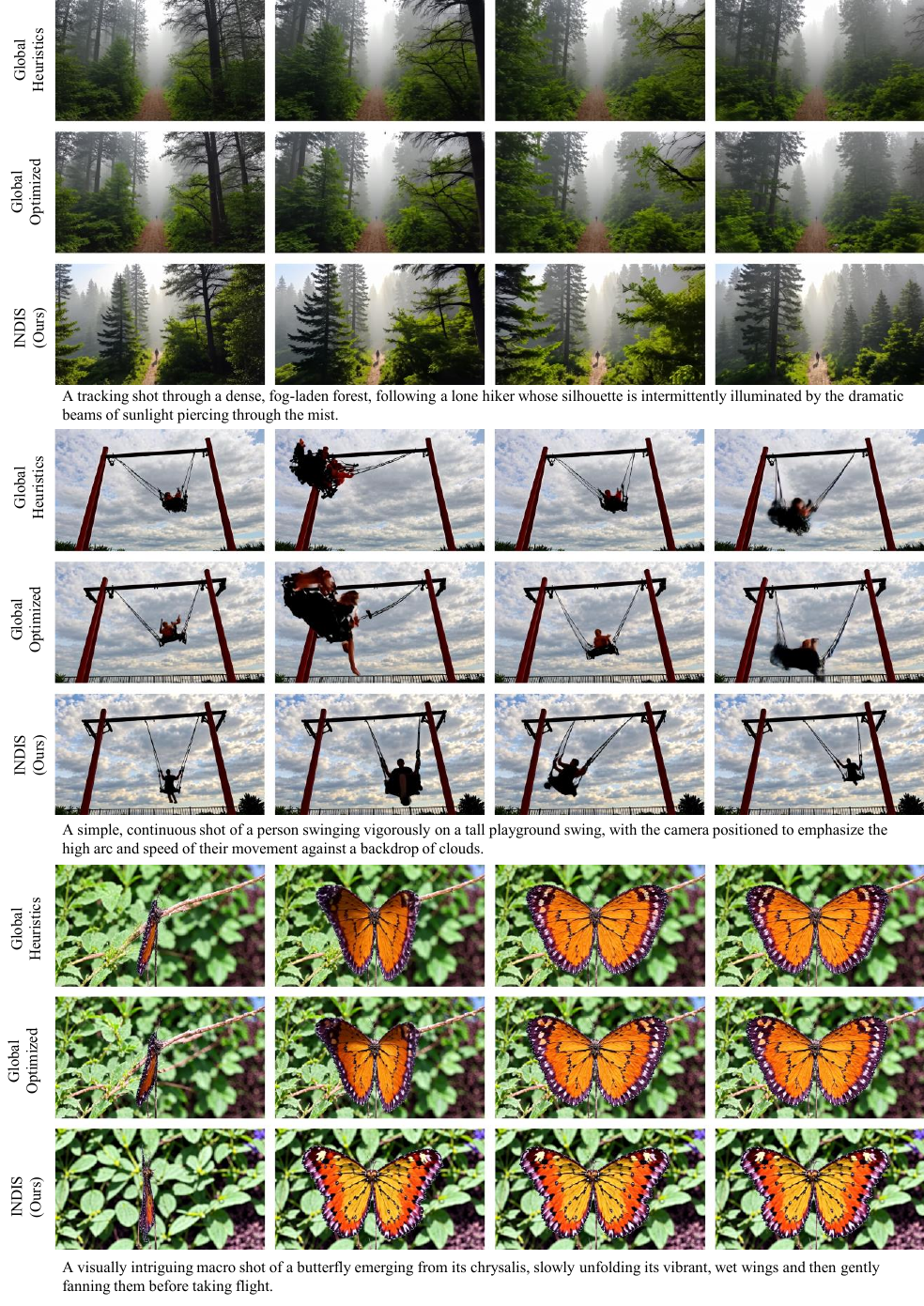}
    \caption{Qualitative Comparison (NFE=5) between global heuristics, global optimized and instance level \(\Ours\) methods (1/2) on LTX-Video}
    \label{figs:ltx_1}
\end{figure*}

\begin{figure*}[b]
    \centering
    \includegraphics[width=0.85\linewidth]{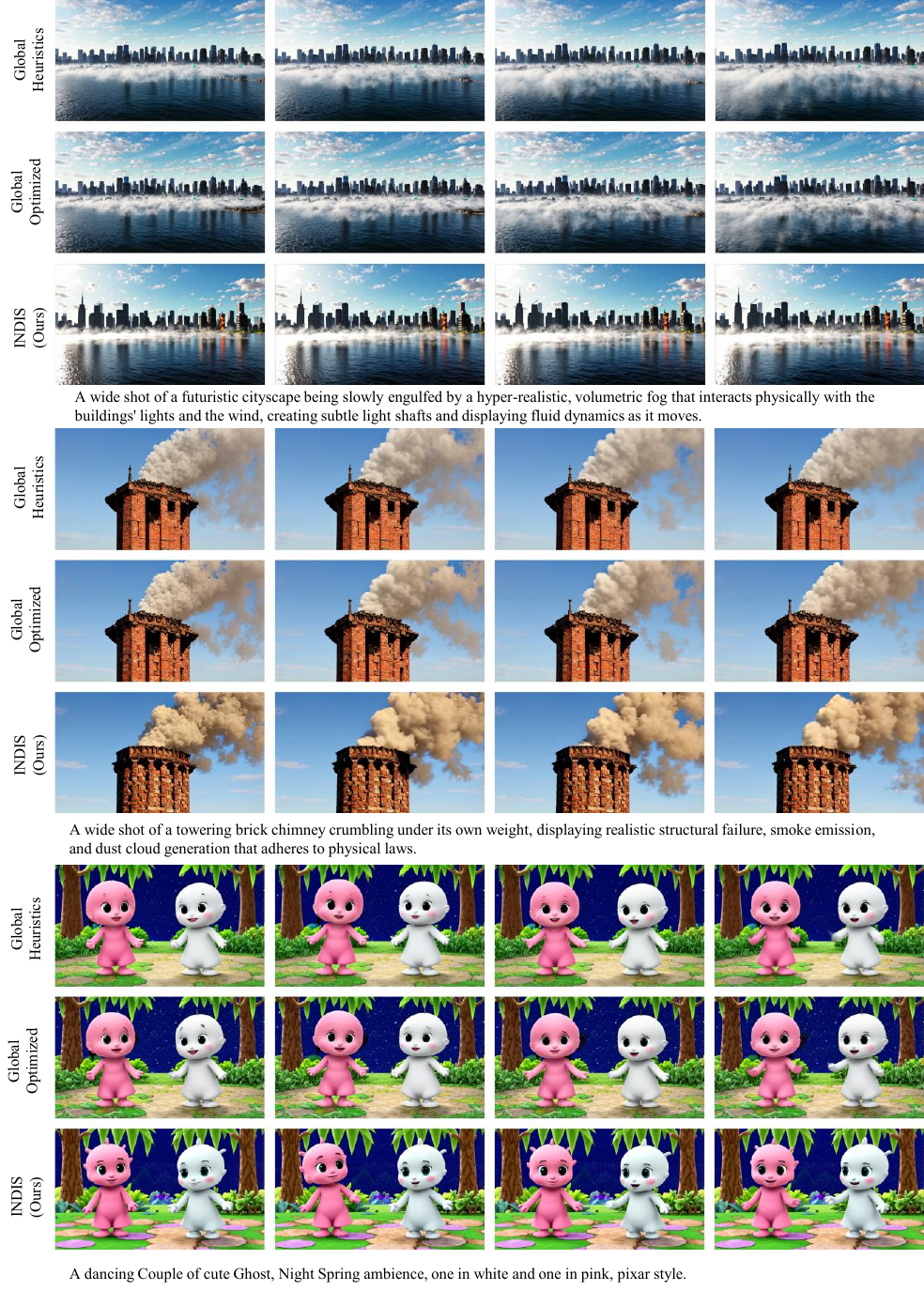}
    \caption{Qualitative Comparison (NFE=5) between global heuristics, global optimized and instance level \(\Ours\) methods (2/2) on LTX-Video}
    \label{figs:ltx_2}
\end{figure*}

\begin{figure*}[b] 
    \centering     

    \begin{subfigure}{\textwidth} 
        \centering
        \includegraphics[width=0.8\textwidth]{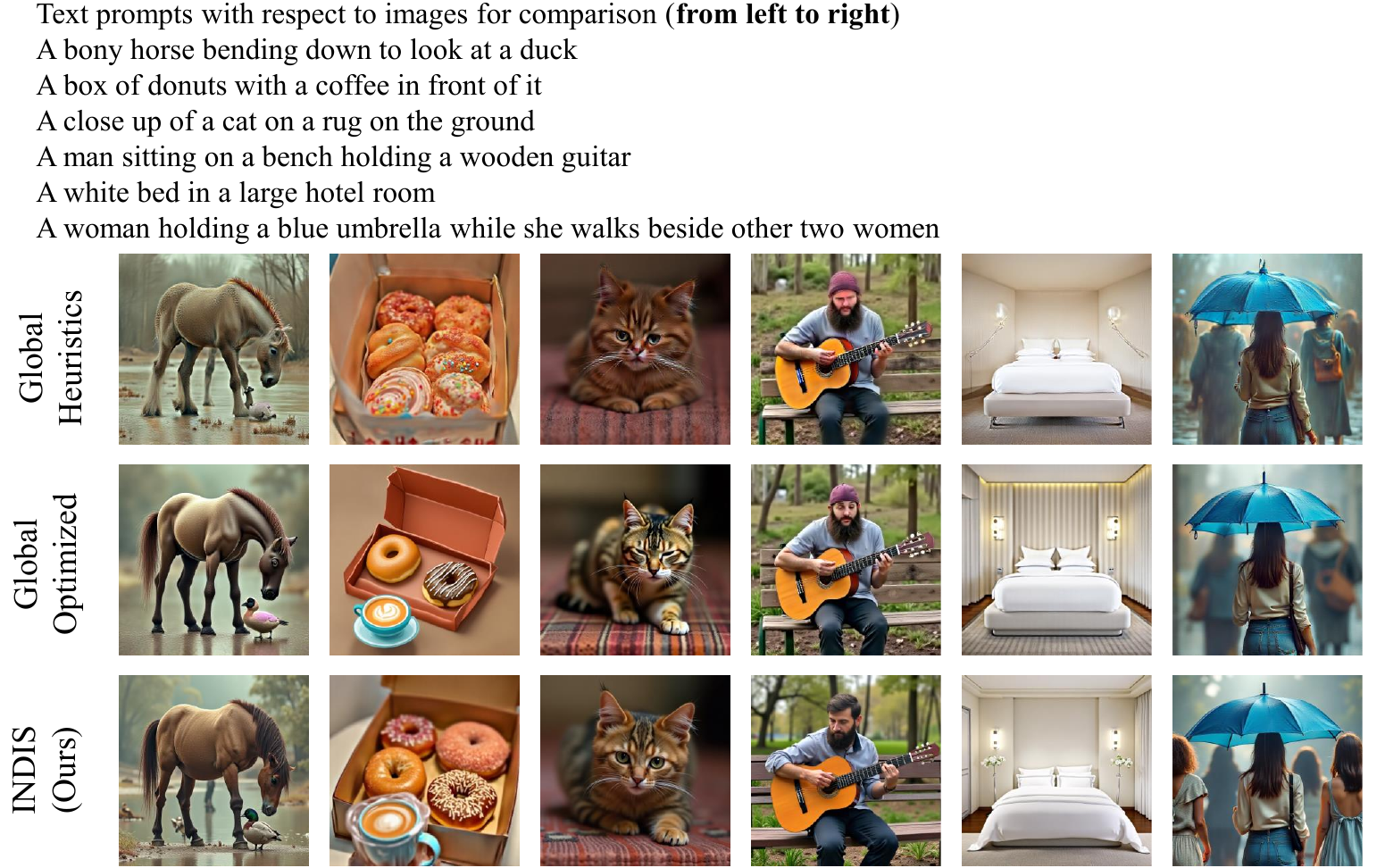} 
        \caption{} 
        \label{fig:nfe6_ours}
    \end{subfigure}

    \vspace{1em} 

    \begin{subfigure}{\textwidth} 
        \centering
        \includegraphics[width=0.8\textwidth]{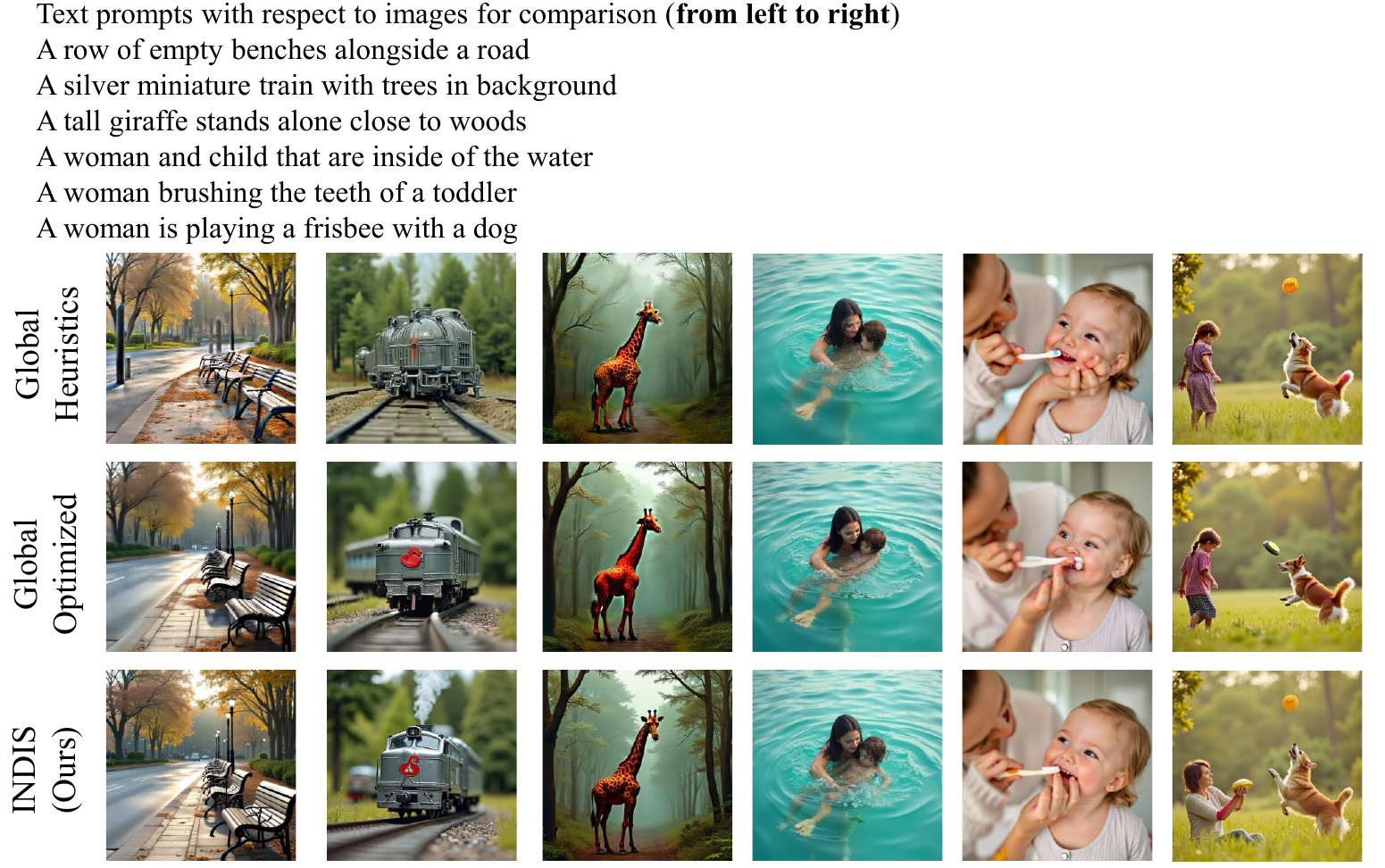} 
        \caption{} 
        \label{fig:nfe6_ours_shifted}
    \end{subfigure}

    \caption{Qualitative Comparison (NFE=7) between global heuristics, global optimized and instance level \(\Ours\) methods (1/2) on FLUX.1-dev}
    \label{fig:qualitative_comparison_nfe6_stacked}
\end{figure*}

\begin{figure*}[p] 
    \centering

    \begin{subfigure}{\textwidth}
        \centering
        \includegraphics[width=0.8\textwidth]{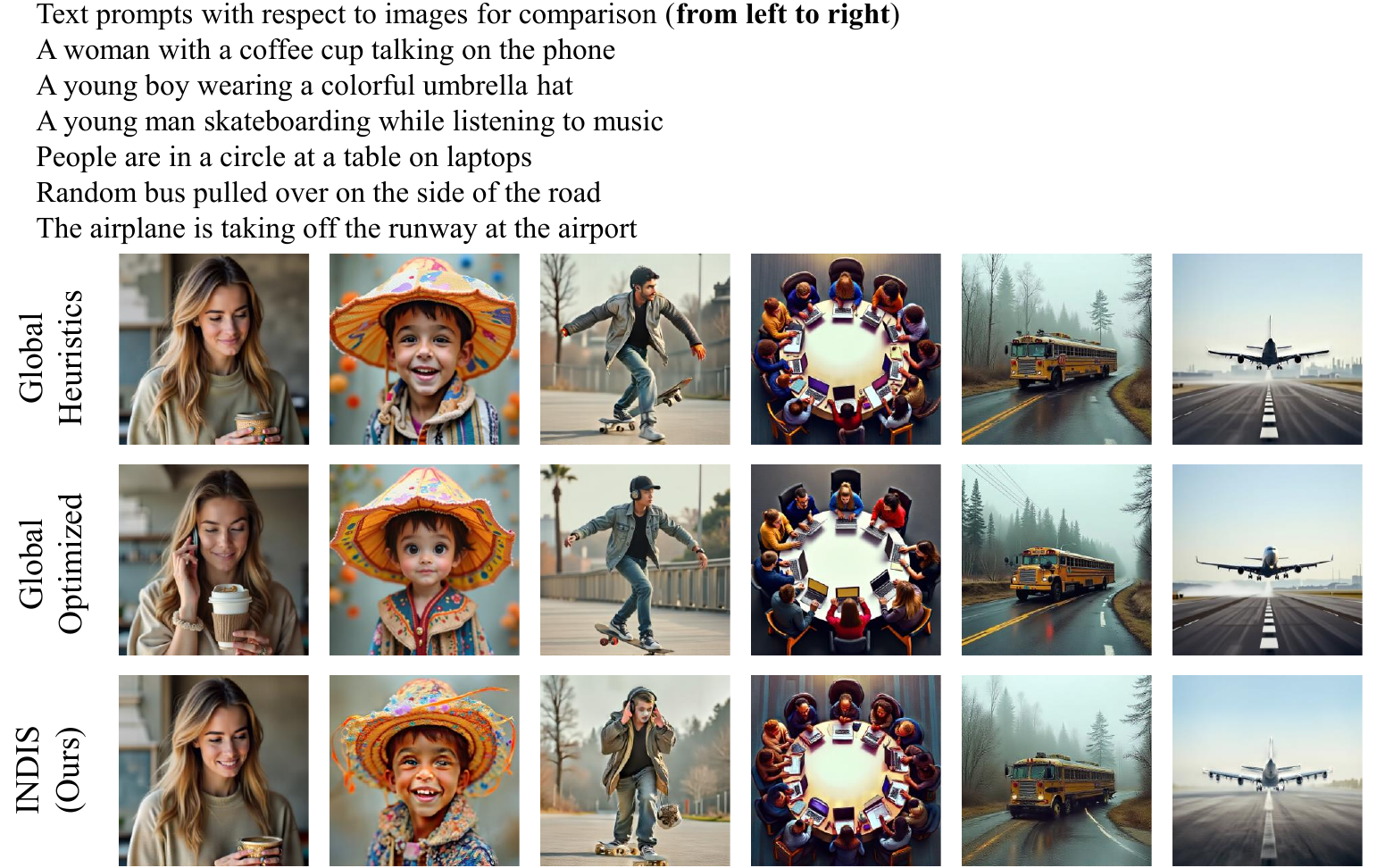} 
        \caption{}
        \label{fig:nfe7_1}
    \end{subfigure}

    \vspace{1em} 

    \begin{subfigure}{\textwidth}
        \centering
        \includegraphics[width=0.8\textwidth]{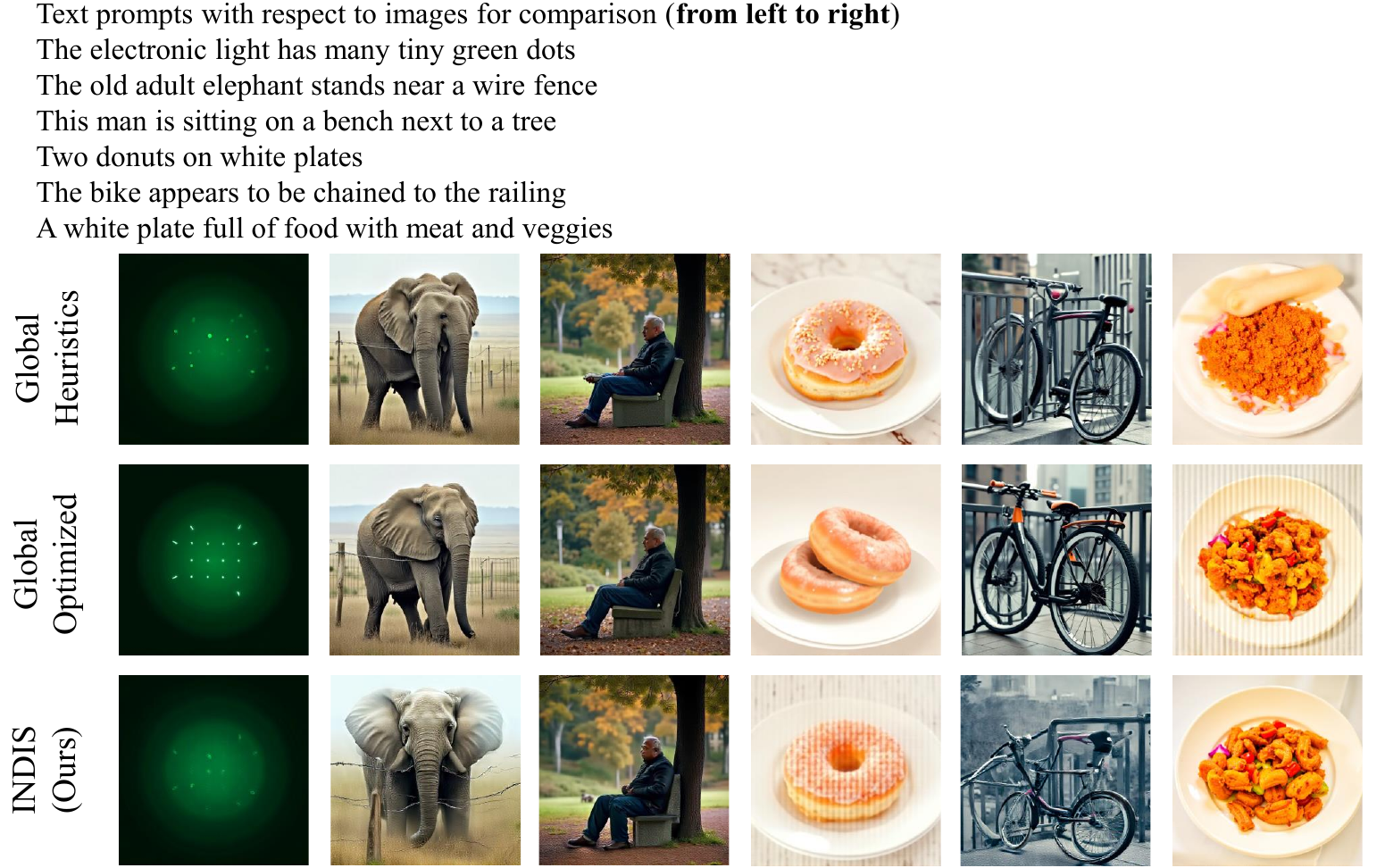} 
        \caption{}
        \label{fig:nfe7_ours_shifted}
    \end{subfigure}

    \caption{Qualitative Comparison (NFE=7) between global heuristics, global optimized and instance level \(\Ours\) methods (2/2) on FLUX.1-dev}
    \label{fig:qualitative_comparison_nfe7_stacked}
\end{figure*}

\begin{figure}[htbp]
    \centering 

    \begin{subfigure}[b]{0.48\textwidth} 
        \centering
        \includegraphics[width=\linewidth]{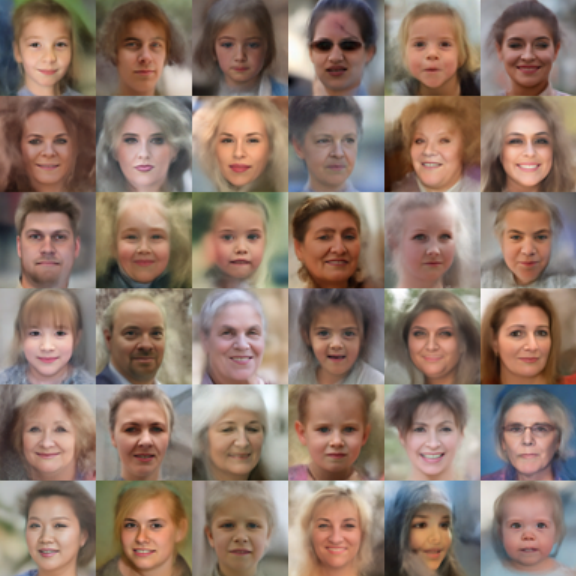}
        \caption{Selected best heuristics.}
        \label{fig:subfig_ffhq} 
    \end{subfigure}
    \hfill 
    \begin{subfigure}[b]{0.48\textwidth} 
        \centering
        \includegraphics[width=\linewidth]{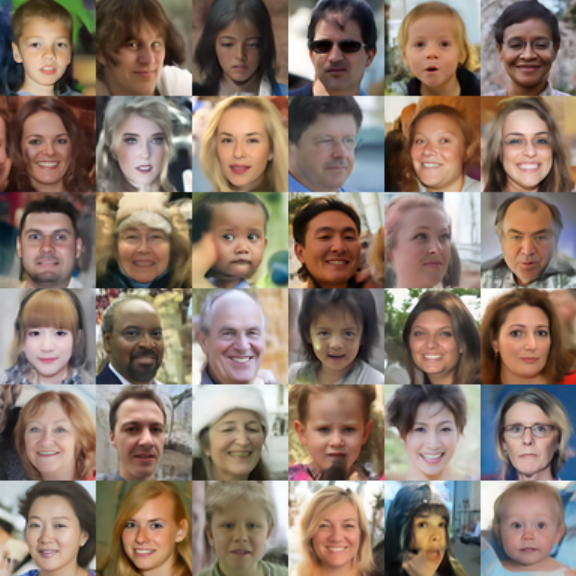}
        \caption{\(\Ours\)}
        \label{fig:subfig_b1} 
    \end{subfigure}
    \caption{Qualitative comparison on FFHQ64x64 datasets with NFE=3 settings.}
\end{figure}

\begin{figure}[htbp]
    \centering 

    \begin{subfigure}[b]{0.48\textwidth} 
        \centering
        \includegraphics[width=\linewidth]{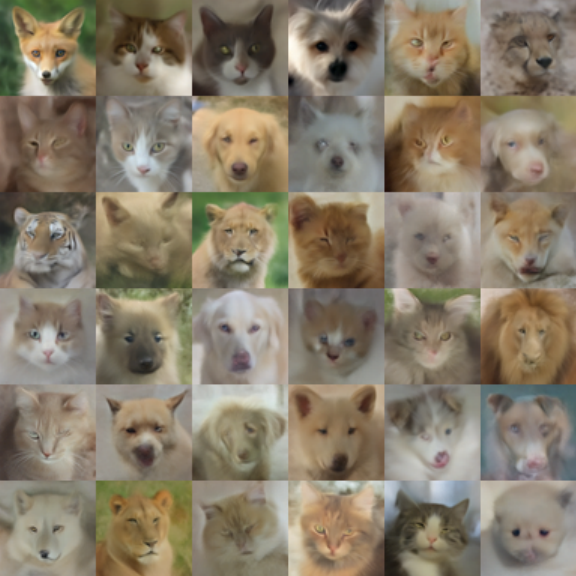}
        \caption{Selected best heuristics.}
        \label{fig:subfig_afhqv2} 
    \end{subfigure}
    \hfill 
    \begin{subfigure}[b]{0.48\textwidth} 
        \centering
        \includegraphics[width=\linewidth]{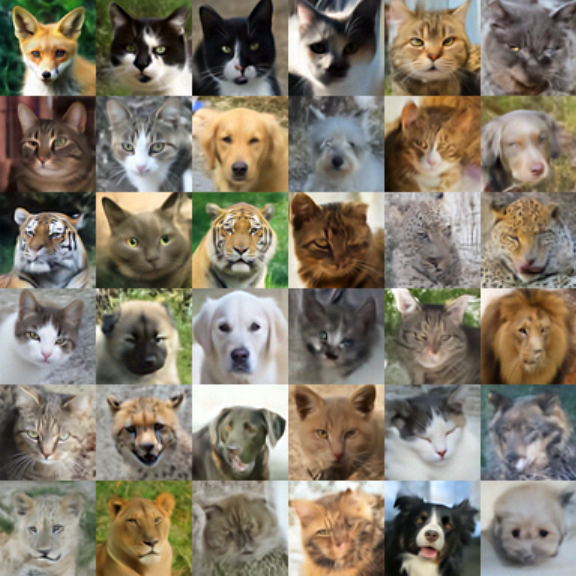}
        \caption{\(\Ours\)}
        \label{fig:subfig_b2} 
    \end{subfigure}
    \caption{Qualitative comparison on AFHQv2 64x64 datasets with NFE=3 settings.}
\end{figure}

\begin{figure}[htbp]
    \centering 

    \begin{subfigure}[b]{0.48\textwidth} 
        \centering
        \includegraphics[width=\linewidth]{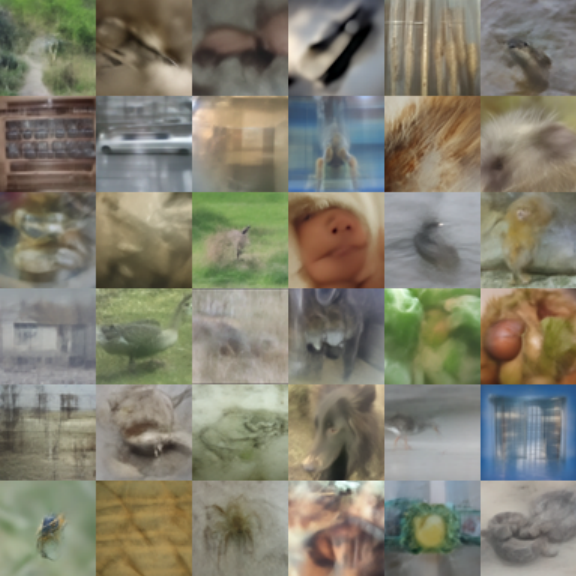}
        \caption{Selected best heuristics.}
        \label{fig:subfig_imagenet64} 
    \end{subfigure}
    \hfill 
    \begin{subfigure}[b]{0.48\textwidth} 
        \centering
        \includegraphics[width=\linewidth]{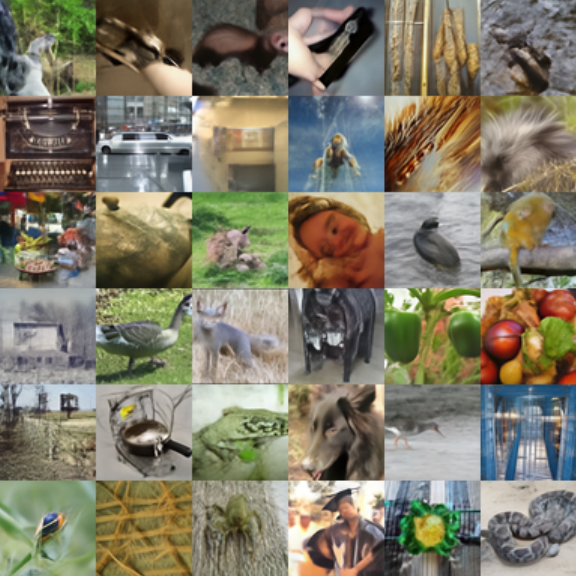}
        \caption{\(\Ours\)}
        \label{fig:subfig_b3} 
    \end{subfigure}
    \caption{Qualitative comparison on ImageNet 64x64 datasets with NFE=3 settings.}
\end{figure}

\begin{figure}[htbp]
    \centering 
    \begin{subfigure}[b]{0.48\textwidth} 
        \centering
        \includegraphics[width=\linewidth]{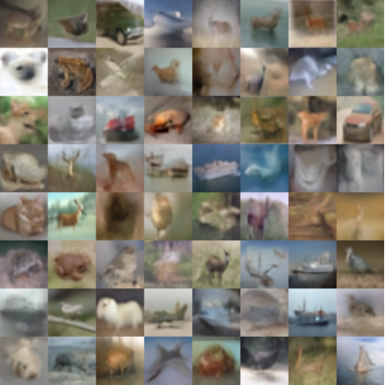}
        \caption{Selected best heuristics.}
        \label{fig:subfig_cifar10} 
    \end{subfigure}
    \hfill 
    \begin{subfigure}[b]{0.48\textwidth} 
        \centering
        \includegraphics[width=\linewidth]{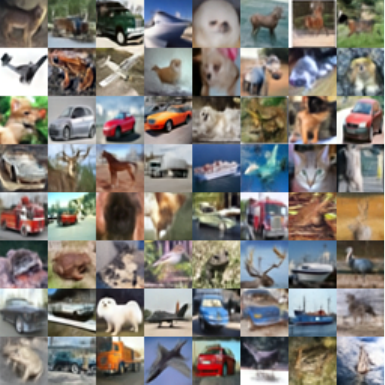}
        \caption{\(\Ours\)}
        \label{fig:subfig_b4} 
    \end{subfigure}
    \caption{Qualitative comparison on CIFAR10 32x32 datasets with NFE=3 settings.}
\end{figure}

\begin{figure}[htbp]
    \centering 

    \begin{subfigure}[b]{0.48\textwidth} 
        \centering
        \includegraphics[width=\linewidth]{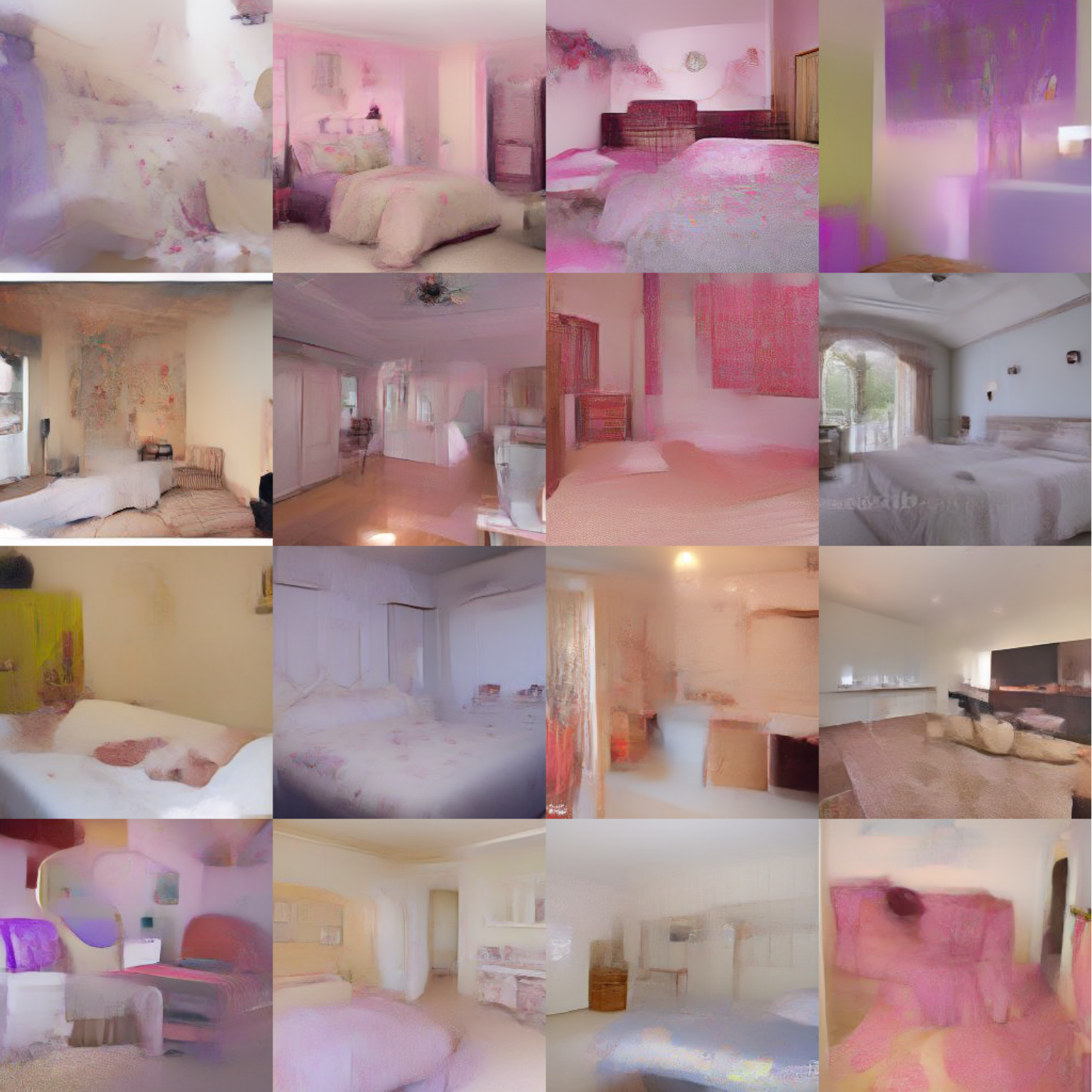}
        \caption{Selected best heuristics.}
        \label{fig:subfig_lsun1} 
    \end{subfigure}
    \hfill 
    \begin{subfigure}[b]{0.48\textwidth} 
        \centering
        \includegraphics[width=\linewidth]{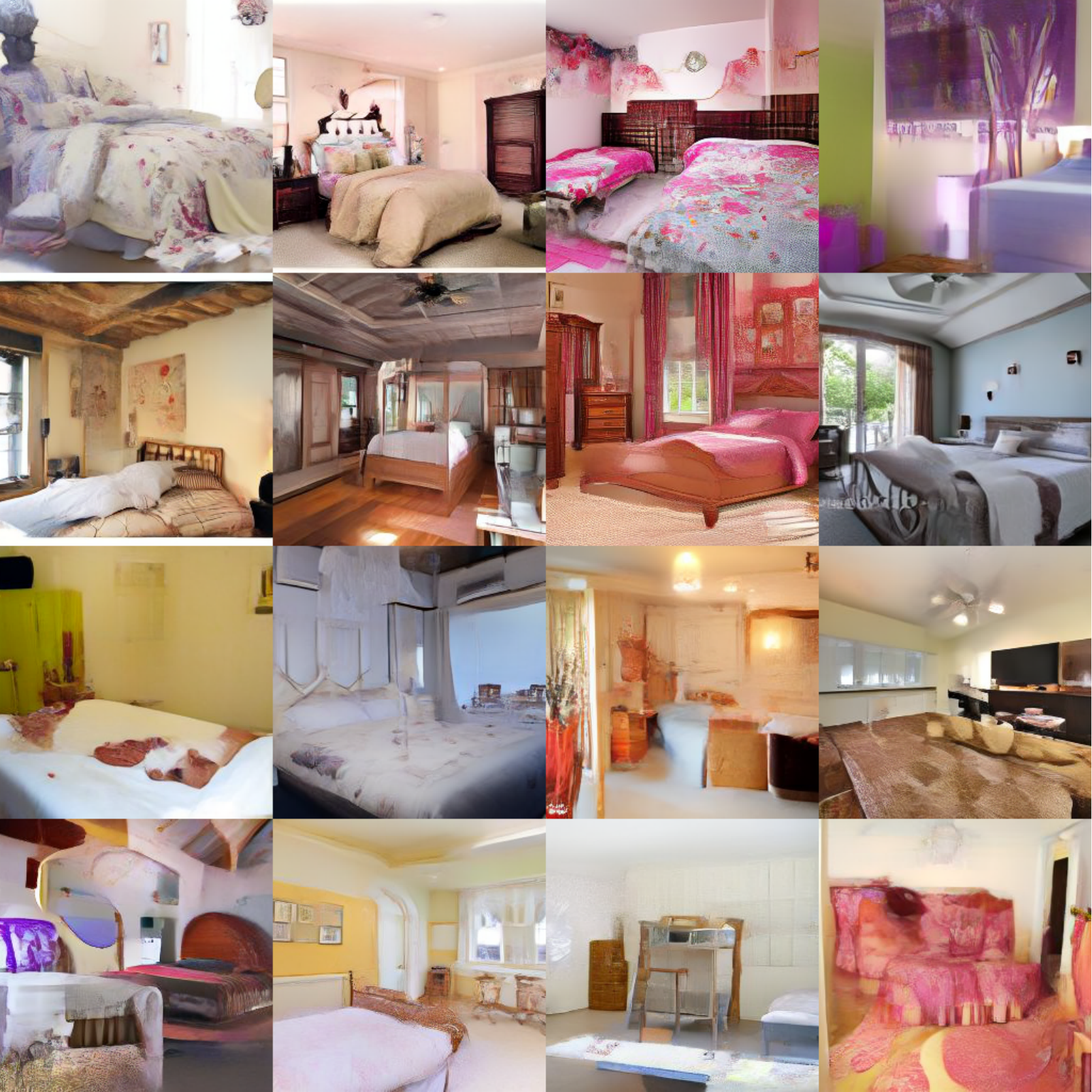}
        \caption{\(\Ours\)}
        \label{fig:subfig_b5} 
    \end{subfigure}
    \caption{Qualitative comparison on latent space LSUN-Bedroom 256x256 datasets with NFE=3 settings.}
\end{figure}

\begin{figure}[htbp]
    \centering 

    \begin{subfigure}[b]{0.48\textwidth} 
        \centering
        \includegraphics[width=\linewidth]{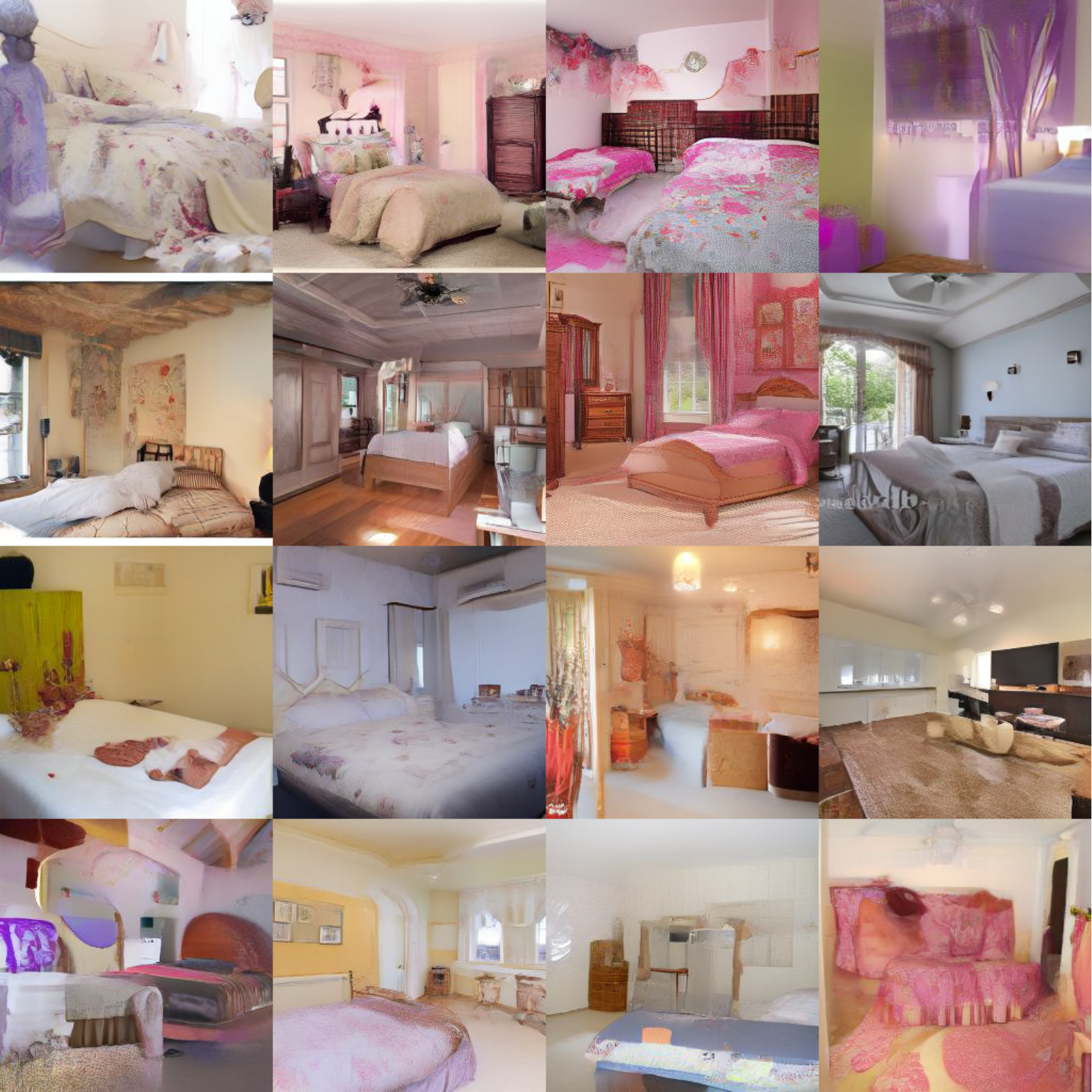}
        \caption{Selected best heuristics.}
        \label{fig:subfig_lsun2} 
    \end{subfigure}
    \hfill 
    \begin{subfigure}[b]{0.48\textwidth} 
        \centering
        \includegraphics[width=\linewidth]{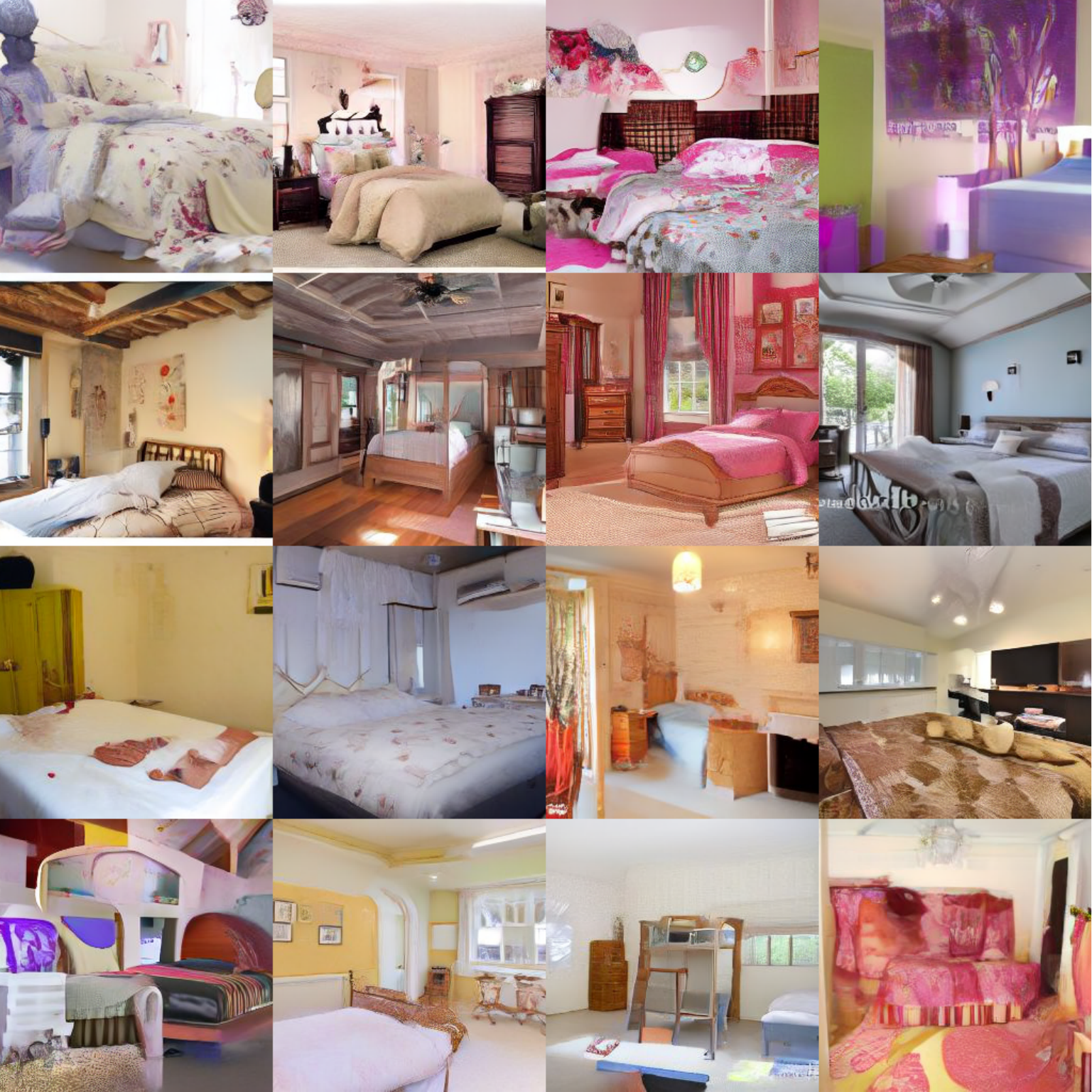}
        \caption{\(\Ours\)}
        \label{fig:subfig_b6} 
    \end{subfigure}
    \caption{Qualitative comparison on latent space LSUN-Bedroom 256x256 datasets with NFE=4 settings.}
\end{figure}

Here we present the qualitative comparison with teacher on FLUX (\Cref{fig:indis_teacher}). Standard comparison against global heuristics and globally optimized baselines on LTX-Video (\Cref{figs:ltx_1,figs:ltx_2}), FLUX.1-dev (512x512) (\Cref{fig:qualitative_comparison_nfe6_stacked,fig:qualitative_comparison_nfe7_stacked}), LSUN-Bedroom (256x256) (\Cref{fig:subfig_lsun1,fig:subfig_lsun2}), CIFAR10 (32x32) (\Cref{fig:subfig_cifar10}), ImageNet (64x64) (\Cref{fig:subfig_imagenet64}), FFHQ (64x64) (\Cref{fig:subfig_ffhq}) and AFHQv2 (64x64) (\Cref{fig:subfig_afhqv2}).


\clearpage
\newpage

\ifarxiv
\else
  \clearpage
  \bibliographystyle{unsrtnat}
\bibliography{ref}
\fi

 \fi

\end{document}